\def\year{2022}\relax
\documentclass[letterpaper]{article} 
\usepackage{bbding}
\usepackage{pifont}
\usepackage{xcolor}
\usepackage{colortbl}
\usepackage{mathtools}
\usepackage{mathrsfs}
\usepackage{multirow}
\usepackage{algorithmic}
\usepackage{subfigure}
\usepackage{array}
\usepackage{tabu}
\usepackage{adjustbox}
\usepackage{tabularx}
\usepackage{booktabs}
\usepackage{diagbox}
\usepackage{array}
\usepackage{stfloats}
\usepackage{float}
\usepackage{caption}
\usepackage{amsthm,amsmath,amssymb}
\usepackage{mathrsfs}
\definecolor{mygray}{gray}{.9}
\newcommand{\textBC}[2]{\textbf{\textcolor{#1}{#2}}}

\usepackage{aaai22}  
\usepackage{times}  
\usepackage{helvet}  
\usepackage{courier}  
\usepackage[hyphens]{url}  
\usepackage{graphicx} 
\usepackage{natbib}  
\usepackage{caption} 
\DeclareCaptionStyle{ruled}{labelfont=normalfont,labelsep=colon,strut=off} 
\usepackage{algorithm}
\usepackage{algorithmic}

\usepackage{newfloat}
\usepackage{listings}
\floatstyle{ruled}
\newfloat{listing}{tb}{lst}{}
\floatname{listing}{Listing}
\title{Self-Supervised Pretraining for RGB-D Salient Object Detection}
\author{
 Xiaoqi Zhao,\textsuperscript{\rm 1}
Youwei Pang, \textsuperscript{\rm 1}
Lihe Zhang, \textsuperscript{\rm 1}\thanks{Corresponding author.}
Huchuan Lu, \textsuperscript{\rm 1,2}
Xiang Ruan \textsuperscript{\rm 3}
}
\affiliations{
   \textsuperscript{\rm 1} Dalian University of Technology, China\\
\textsuperscript{\rm 2} Peng Cheng Laboratory, China\\
\textsuperscript{\rm 3} Tiwaki Co.,Ltd., Japan\\
	
\{zxq,lartpang\}@mail.dlut.edu.cn, \{zhanglihe,lhchuan\}@dlut.edu.cn,\ 
ruanxiang@tiwaki.com
}

\usepackage{bibentry}

\begin{document}

\maketitle

\begin{abstract}
Existing CNNs-Based RGB-D salient object detection (SOD) networks are all required to be pretrained on the ImageNet to learn the hierarchy features which helps  provide a good initialization. However, the collection and annotation of large-scale datasets are time-consuming and expensive. In this paper, we utilize self-supervised representation learning (SSL) to design two pretext tasks: the cross-modal auto-encoder and the depth-contour estimation. Our pretext tasks require only a few and unlabeled RGB-D datasets to perform pretraining, which makes the network capture rich semantic contexts and reduce the gap between two modalities, thereby providing an effective initialization for the downstream task. In addition, for the inherent problem of cross-modal fusion in RGB-D SOD, we propose a consistency-difference aggregation (CDA) module that splits a single feature fusion into multi-path fusion to achieve an adequate perception of consistent and differential information.
The CDA module is general and suitable for cross-modal and cross-level feature fusion. 
Extensive experiments on six benchmark datasets show that our self-supervised pretrained model performs favorably against most state-of-the-art methods pretrained on ImageNet.  The source code will be publicly available at 
\textcolor{red}{\url{https://github.com/Xiaoqi-Zhao-DLUT/SSLSOD}}.
\end{abstract}

\section{Introduction}
RGB-D salient object detection (SOD) task aims to utilize the depth map, which contains stable geometric structures and extra contrast cues, to provide important supplemental information for handling complex environments such as low-contrast salient objects that share similar appearances to the background. Benefiting from Microsoft Kinect, Intel RealSense, and some modern smartphones (e.g., Huawei Mate30, iPhone X, and Samsung Galaxy S20), depth information can be conveniently obtained. 

With the development of deep convolutional neural networks (CNNs), many CNNs-Based SOD methods~\cite{zhang2019memory,zhang2020LFNet,li2021joint,LSSA,BBSNet,CMWNet,UCNet,DFMNet,DSA2F,MSNet,MSAPS,HAINet,GateNet,MINet,ji2021learning,YOFO,zhang2021dynamic,HDFNet} can achieve satisfactory performance. They all are required to be pretrained on the ImageNet~\cite{ImageNet} to learn rich and high-performance visual representations for downstream tasks. 
However, the ImageNet contains about $1.3$ million labeled images covering 1,000 classes while each image is labeled by human workers with one class label. Such expensive labor costs are immeasurable.
Recently, Doersch \textit{et al.}~\cite{UVRL-context}, Wang and Gupta~\cite{UVRL-videos} and Agrawal \textit{et al.}~\cite{LSM} have explored a novel paradigm for unsupervised learning called self-supervised learning (SSL). The main idea is to exploit different labeling that is freely available besides or within visual data and to use them as intrinsic reward signals to learn general-purpose features. In the learning process, the context has proven to be a powerful source of automatic supervisory signal for learning representations~\cite{context1,context2,context3,context4}. The SSL needs to design a ``pretext'' task to learn rich context and then utilizes the pretrained model for some other ``downstream'' tasks, such as classification, detection, and semantic segmentation. 
In addition, for the RGB-D SOD task, how to fully integrate the features of two modalities is still an open problem of great concern, just like the cross-level feature fusion.
How to better design a general module suitable for feature pairs with multiple complementary relationships is a currently neglected issue.

Two-stream RGB-D SOD networks~\cite{DMRA,LSSA,JLDCF,HDFNet,CDNet,CoNet} usually load ImageNet-pretrained weights. Their encoders for RGB and depth streams have the same pretraining task, i.e., image classification. Their decoders also have the same task of saliency prediction. This kind of task homogeneousness in both encoder and decoder can greatly reduce the inter-modal gap between the two streams.
We know that image classification network often activates the corresponding semantic region in feature maps (e.g., Class Activation Map), while the depth map possibly highlights salient regions. Inspired by this, we firstly design a pretext task of depth estimation, which can promote the RGB encoder to capture localization, boundary and shape information of objects by comparing relative spatial positions at the pixel level. Moreover, the depth map plays an information filtering and attention role in RGB-D SOD, which is the same as the class activation map in image classification. Secondly, in order to reinforce the cross-modal information interaction, we design another pretext task of reconstructing RGB channels from the depth map. This task requires that the network learns the way to assign the colors for different positions. Due to the limited prior information, this is a very difficult task, which can stimulate the potential of representation learning and drive the depth encoder to capture the cues of shapes and semantic relationships among different objects and fore/backgrounds.
The above two pretext tasks actually form a cross-modal auto-encoder, which only needs pairs of RGB and Depth images without any mutual labels. 

In the decoder, we design a pretext task of depth-contour estimation. The reason is twofold: (1) Similar prediction tasks help reduce the gap between the modalities. The contour prediction is an analogous process to saliency detection. 
(2) The depth contour is cleaner than the RGB contour and tends to better depict edge information about salient object/objects, since based on human cognition, salient objects usually have more pronounced depth difference from the background. Moreover, the contour is also an important attribute of salient object. Once the pretext can predict the contour of the salient object well, it simplifies the downstream RGB-D SOD task to predict fore/background properties inside and outside the contour. After these pretext networks have been trained, the RGB-D SOD network can gain good initialization. 

About the network architecture, we propose a general module called consistency-difference aggregation (CDA) to achieve both cross-modal and cross-level fusion. Specifically, for two types of features with complementary relationships, we calculate their jointly consistent (JC) features and jointly differential (JD) features. The JC  maintains more attention to their consistency and suppresses the interference of non-salient information, while the JD depicts their difference in salient region and encourages the cross-modal or cross-level alignment. By saliency guided consistency-difference aggregation, the gap between modalities or levels is greatly narrowed. 

Our main contributions can be summarized as follows:
\begin{itemize}
     \item We present a self-supervised network closely related to the RGB-D SOD task, which consists of a cross-modal auto-encoder and a depth-contour estimation decoder. It is the first method to use self-supervised representation learning for RGB-D SOD.
     
    \item We design a simple yet effective consistency-difference aggregation structure that is suitable for both cross-level and cross-modal feature integration.
   
    \item We use $6,392$ pairs of RGB-Depth images (without manual annotations) to pretrain the model instead of using ImageNet ($1,280,000$ with image-level labels). Our model still performs much better than most competitors on six RGB-D SOD datasets.
\end{itemize}

\begin{figure*}
    \includegraphics[width=0.90\textwidth]{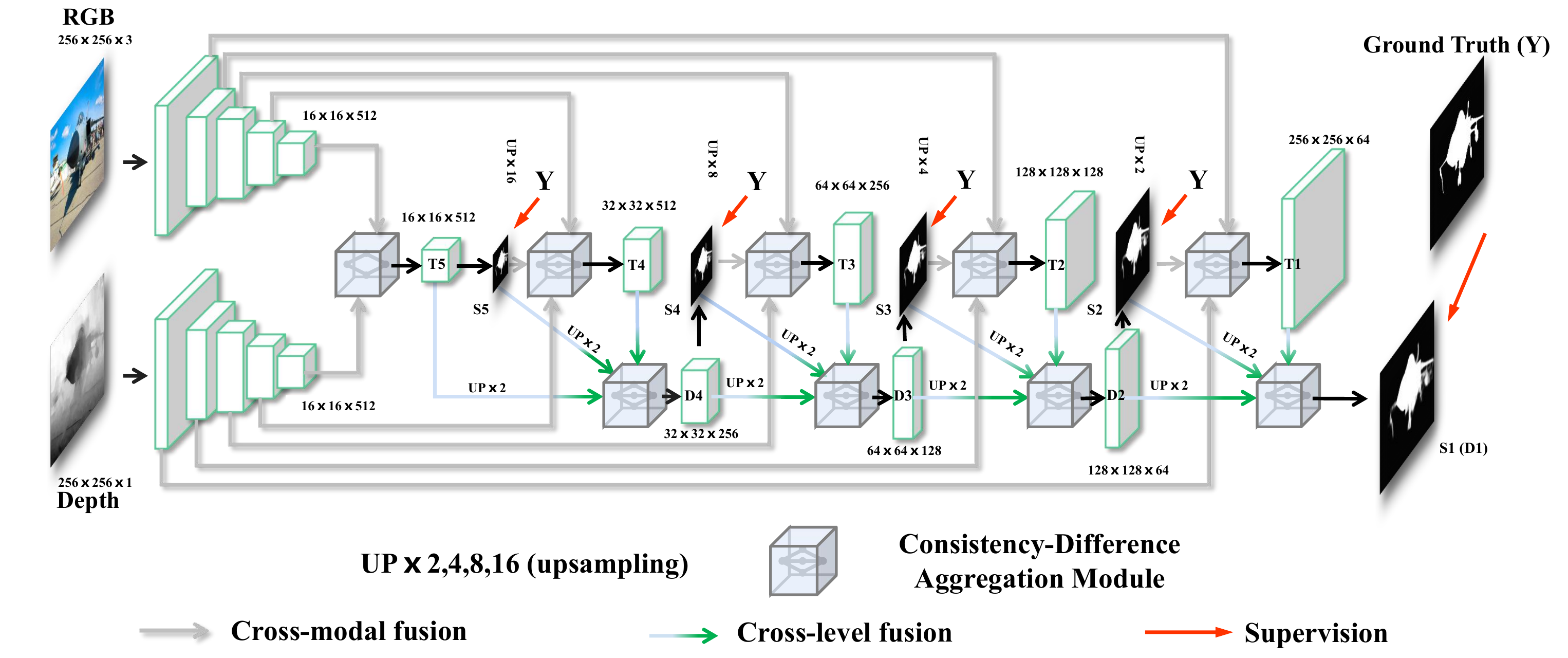}\\
    \centering
     \setlength{\abovecaptionskip}{-10pt}
    \caption{Network pipeline of the downstream task. It consists of two VGG-16 encoders, five cross-modal layers and four decoder blocks. Both the cross-modal and cross-level fusions are achieved by the consistency-difference aggregation (CDA) module. We generate ground truth of multiple resolutions and use cross-entropy loss as supervision.} 
    \label{fig:Figure3}
      \vspace{-5.5mm}
\end{figure*} 

\section{Related Work}
\subsection{RGB-D Salient Object Detection}
Generally speaking,  the depth map can be utilized in three ways: early fusion~\cite{early_fusion_1,early_fusion_2,DANet}, middle fusion~\cite{Middle_fusion} and late fusion~\cite{late_fusion}. According to the number of encoding streams, RGB-D SOD methods can divided into two-stream~\cite{CPFP,PGAR,BBSNet,LSSA,UCNet,DFMNet,DSA2F,DCF} and single-stream~\cite{DANet} ones. The two-stream networks mainly focus on sufficiently incorporating cross-modal complementarity.  Liu \textit{et al.}~\cite{LSSA} utilize the self-attention and the other modality’s attention in the Non-local structure to fuse multi-modal information.  Chen and Fu ~\cite{PGAR} propose an alternate refinement strategy and combine a guided residual block to generate both refined feature and refined prediction.  Sun \textit{et al.}~\cite{DSA2F} design a new NAS-based model for the heterogeneous feature fusion in RGB-D SOD. These two-stream designs significantly increase the number of parameters in the network. Recently, Zhao \textit{et al.}~\cite{DANet} combine depth map and  RGB image from starting to build a real single-stream network to take advantage of the potential contrast information provided by the depth map, which provides a new perspective on the RGB-D SOD field. Besides, different level features have different characteristics. High-level ones have more semantic information which helps localize the objects, while low-level ones have more detailed information which can capture the subtle structures of objects. 
However, both two-stream and single-stream networks belittle the cross-level fusion, which may lead to a significant reduction in the effectiveness of the cross-modal fusion. In this work, we propose a consistency-difference aggregation module, which can be applied in universal combination of multiple complementary relationships (RGB/Depth, High/Low level).

\subsection{Self-Supervised Representation Learning}
Self-supervised learning (SSL) is an important branch of unsupervised learning technique. It refers to the learning paradigm in which ConvNets are explicitly trained with automatically generated labels. During the training phase, a predefined pretext task is designed for ConvNets, and the pseudo labels in the pretext task are automatically generated based on some attributes of data. Then the ConvNet is trained to learn object functions of the pretext task. After the SSL training is finished, the learned weights are transferred to downstream tasks as their pretrained models. This strategy can overcome overfitting of small sample problem and obtain the generalization capability of ConvNets.

Many pretext tasks have been designed and applied for self-supervised learning, such as image inpainting~\cite{image_inpainting_ssl}, clustering~\cite{cluster_ssl}, image colorization~\cite{image_colorization_ssl}, temporal order verification~\cite{temporal_order_verification_ssl} and visual audio correspondence verification~\cite{visual_audio_correspondences_verification_ssl}. 
Effective pretext tasks can promote ConvNets to learn useful semantic features for the downstream task. 
In this work, we rely on the characteristics of RGB-D SOD task to design two related pretext tasks: cross-modal auto-encoder and depth-contour estimation. The former encourages the two-stream encoder to learn each modal information and reduce the inter-modal gap. The latter can further facilitate the cross-modal fusion, which also provides a good feature prior to ease the capture of object-contour and object-localization.

\section{The Proposed Method}
In this section, we first describe the overall architecture of the proposed RGB-D SOD network. And then, we present the details of the consistency-difference aggregation (CDA) module for cross-modal and cross-level feature fusion. Next, we introduce the designed pretext tasks: cross-modal auto-encoder and depth-contour estimation. Finally, we list all the supervisions and loss functions used in the network for both the pretext tasks and the downstream task.
\subsection{The Overall Architecture}
Our network architecture, shown in Fig.~\ref{fig:Figure3}, follows a two-stream model that consists of five encoder blocks, five transition layers ($\mathbf{T}^i$ $i \in \left \{1, 2, 3, 4, 5 \right \}$ ), four decoder blocks ($\mathbf{D}^i$ $i \in \left \{1, 2, 3, 4\right \}$ ) and nine consistency-difference aggregation modules. The encoder-decoder architecture is based on the FPN~\cite{FPN}. The encoder is based on a common backbone network, e.g.,  VGG-16~\cite{VGG}, to perform feature extraction on RGB and Depth, respectively.  
 We cast away all the fully-connected layers and the last pooling layer of the VGG-16 to modify it into a fully convolutional network. We convey the output features from the encoding blocks of two modalities to the consistency-difference aggregation module to achieve cross-modal fusion at each level. 
 And the CDA is also embedded in the decoder. Once  these cross-modal fused features are obtained, they will participate in the decoder and gradually integrate the details from high-level to low-level, thereby continuously restoring the full-resolution saliency map. 
\begin{figure}[t]
\includegraphics[width=\linewidth]{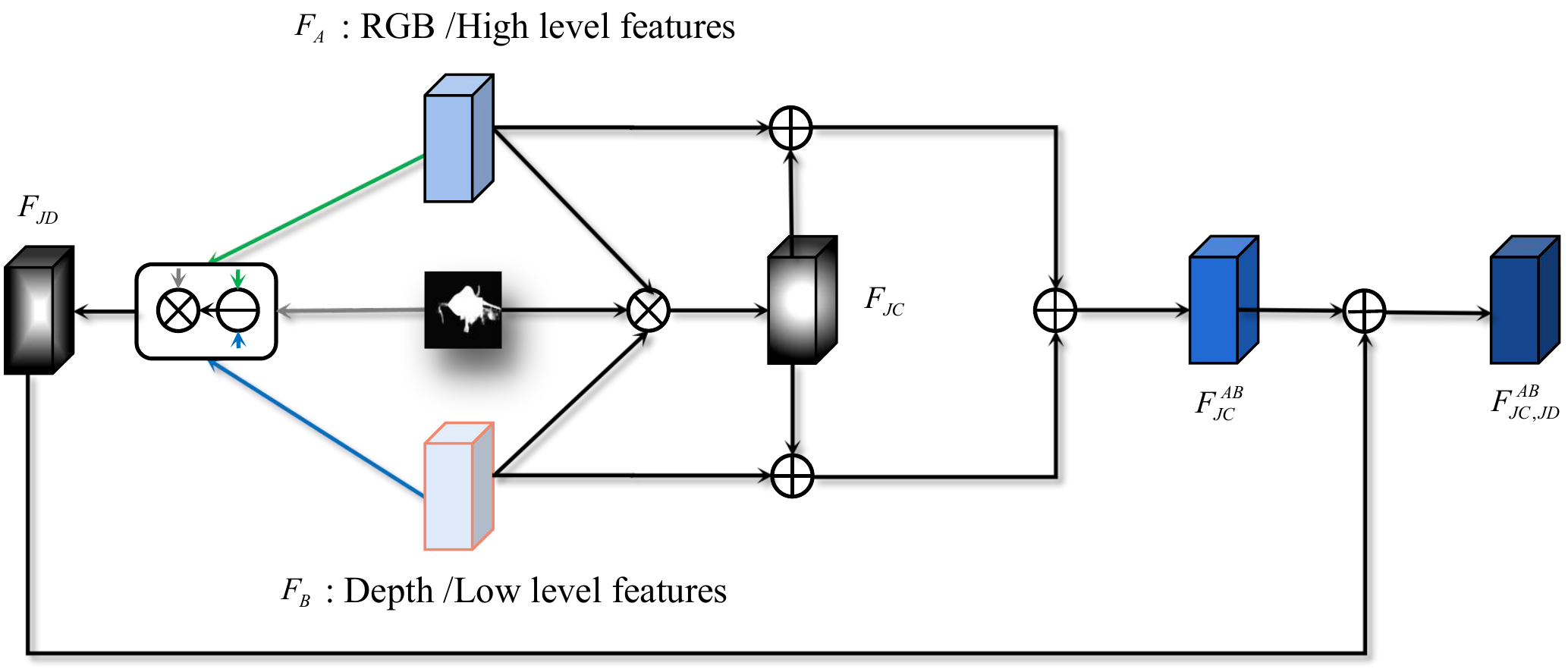}\\
        \centering
         \setlength{\abovecaptionskip}{-10pt}
        \caption{Illustration of the consistency-difference aggregation module.}
\label{fig:Figure4}
\vspace{-5.5mm}
\end{figure} 

\subsection{Consistency-Difference Aggregation Module}\label{sec:MPSF}
The consistency-difference aggregation module can strengthen the consistency of features and highlight their differences, thereby increasing inter-class difference and reducing intra-class difference. Fig.~\ref{fig:Figure4} shows the internal structure of the proposed CDA module. We use  $F_{A}$ and $F_{B}$ to represent different modality or level feature maps. They all have been activated by the ReLU operation. First, we use element-wise multiplication between $F_{A}$ and $F_{B}$ at the same time constrained by saliency map $S$ of the side-out prediction to obtain  high-confidence region features, which are usually consistent with salient objects. This process can be formulated as follows:
 \begin{equation}\label{equ:1}
 \begin{split}
     F_{JC} = Conv(F_{A} \otimes F_{B} \otimes S),
 \end{split}
\end{equation}
where $\otimes$ is the element-wise multiplication and $Conv(\cdot)$ denotes the convolution layer. 
Next, the jointly consistent features are applied to enhance saliency cues in $ F_{A}$ and $F_{B}$, thereby yielding initial fused features,
 \begin{equation}\label{equ:2}
 \begin{split}
     F_{JC}^{AB} = Conv(Conv(F_{JC} \oplus F_{A}) \oplus Conv(F_{JC} \oplus F_{B})),
 \end{split}
\end{equation}
 where $\oplus$ is the element-wise addition.  $F_{JC}^{AB}$ achieves consistency boosting of features, especially in salient regions. We calculate the jointly differential features of $F_{A}$ and $F_{B}$:
  \begin{equation}\label{equ:2}
 \begin{split}
     F_{JD} = Conv(\vert F_{A} \ominus F_{B} \vert \otimes S),
 \end{split}
\end{equation}
 where $\ominus$ is the element-wise subtraction and $ \vert \cdot \vert$ calculates the absolute value. $F_{JD}$ depicts the feature difference in salient regions. The final fusion are generated by combining  $ F_{JC}^{AB}$ and $F_{JD}$: 
 \begin{equation}\label{equ:3}
 \begin{split}
     F_{JC,JD}^{AB} = Conv( F_{JC}^{AB} \oplus F_{JD}).
 \end{split}
\end{equation}
Compared with $F_{JC}^{AB}$, $F_{JC,JD}^{AB}$ contains richer complementary information about salient objects. 
Through a series of addition and subtraction, the features are well aligned under the constraint of the side-out prediction, thereby progressively promoting the correspondence between feature distribution and fore/background categories in the common space, as shown in Fig.~\ref{fig:Figure7}.

\begin{figure}[t]
\includegraphics[width=\linewidth]{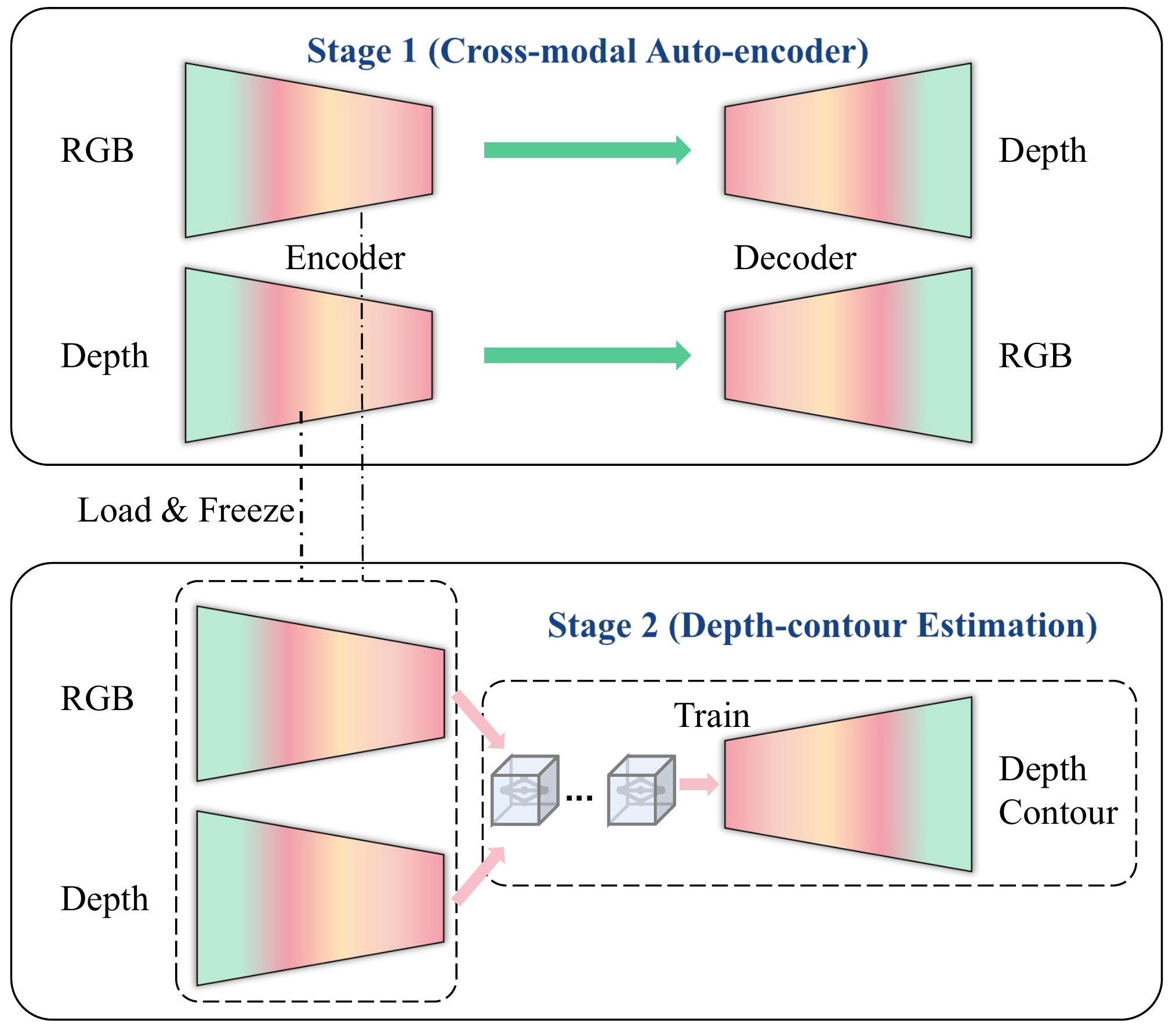}\\
        \centering
         \setlength{\abovecaptionskip}{-10pt}
        \caption{Network pipeline of the pretext tasks. Stage 1: Cross-modal auto-encoder. Stage 2: Depth-contour estimation. The network in Stage 2 is equipped with the  CDAs.}
\label{fig:Figure5}
\vspace{-5mm}
\end{figure} 
\subsection{Pretext Tasks: Cross-modal Auto-encoder and Depth-contour Estimation}
Instead of previous approaches which obtain effective initial representations through ImageNet pretraining, we design a pure self-supervised network to mine RGB-D information without manual annotations. Fig.~\ref{fig:Figure5} shows the structure of the proposed SSL network, which consists of two components, namely, the cross-modal auto-encoder and the depth-contour estimation decoder. 

In the first stage, we use pairs of RGB-Depth data to predict each other, i.e., using the RGB image to predict the depth map and using the depth map to reconstruct the RGB image. The backbone of these two networks is based on VGG-16 with random initialization. We directly adopt the FPN as the basic structure.  Through the cross-modal auto-encoder, the encoder of the RGB or depth stream can respectively turn to capture the information of another modal.
As a result, the features of the two encoders tend to converge into a common space, which greatly reduces the gap between modalities to facilitate subsequent cross-modal fusion.
Moreover, in the process of predicting each other, the contextual representation capabilities can be effectively learned in their own streams. As suggested in~\cite{difficult_ssl}, increasing the complexity of the pretext task is generally beneficial to the performance of the SSL. 
In this work, our cross-modal encoders do not interact with each other in any way. The completely independent cross-modal generation obviously elevates the difficulty, making the network try its best to learn useful semantic cues about the two modalities. 

In the second stage, we load the parameters of the encoders trained in the first stage to calculate multi-scale features in each encoding stream. And then, we exploit the designed CDA module described in Sec.~\ref{sec:MPSF} to achieve the cross-modal and cross-level fusion for predicting the depth-contour. This pretext task allows further cross-modal integration, and the learned features provide contour prior (i.e., a kind of fore/background semantic guidance) for the downstream task to predict saliency map.

\subsection{Supervision}\label{sec:supervision}
In our RGB-D SOD network, the total loss is written as:
\begin{equation}\label{equ:5}
 \begin{split}
    L_{sod}= \sum_{n=1}^{N}({l_{bce}^{w}}^{(n)} + {l_{iou}^{w}}^{(n)}),
 \end{split}
\end{equation}
where $l_{bce}^{w}$ and $l_{iou}^{w}$ represent the weighted IoU loss and binary cross entropy (BCE) loss which have been widely adopted in segmentation tasks. We use the same definitions as in ~\cite{F3Net,PraNet,MSNet}. $N$ denotes the number of the side-out. Our RGB-D SOD model is deeply supervised with five outputs, i.e. $N$ = 5. 

The cross-modal auto-encoder task consists of depth estimation and RGB reconstruction. Since both RGB and depth information presents patch consistency in a scene, we follow the related depth estimation works~\cite{depth3,depth4,depth5} to adopt the combination loss of the L1 and SSIM~\cite{SSIM}. The use of SSIM is well suited as a loss function for the patch-level regularization. The L1-loss calculates the absolute distance between the predicted map and ground truth, which is supervised at the pixel-level. The combination of SSIM-loss and L1-loss allows the network to have both patch-level and pixel-level supervision, thereby promoting more locally consistent prediction.

For the depth-contour estimation task, its ground truth (G$_{dc}$) is calculated based on the depth map (G$_{d}$) provided by the RGB-D datasets. Specifically, we employ the morphological dilation and erosion as follows:
\begin{equation}\label{equ:5}
 \begin{split}
    G_{dc}= D_{m}(G_{d})-E_{m}(G_{d}),
 \end{split}
\end{equation}
where $D(\cdot)$ and $E(\cdot)$ are the dilation and erosion operations, respectively. $m$ denotes a full one filter of size $m\times m$ for erosion and expansion. It is set to $5$ in this paper. We only use the L1-loss for each side-out during the training phase.

\section{Experiments}
\subsection{Datasets}
We evaluate the proposed model on six public RGB-D SOD datasets which are \emph{RGBD135}~\cite{RGBD135}, \emph{DUT-RGBD}~\cite{DMRA}, 
\emph{STERE}~\cite{STERE}, \emph{NLPR}~\cite{early_fusion_1}, \emph{NJUD}~\cite{NJU2000} and \emph{SIP}~\cite{SIP}.
%
For the downstream task, we adopt the same training set as most methods~\cite{A2DELE,DANet,CMWNet}, that is, $800$ samples from the DUT-RGBD, $1,485$ samples from the NJUD and $700$ samples from the NLPR are used for training. The remaining images and other three datasets are used for testing. For the pretext tasks, we combine  training subsets of NJUD and NLPR with the recent  DUTLF-V2~\cite{DUTLF-V2}  ($4,207$ samples) to finish pretraining, that is, $6,392$ images in total.
\subsection{Evaluation Metrics}
We adopt several widely used metrics for quantitative evaluation: F-measure ($F_{\beta}^{max}$)~\cite{colorcontrast_Fm}, the weighted F-measure ($F_{\beta}^{w}$)~\cite{Fwb}, mean absolute error (MAE, $\mathcal{M}$), the recently released S-measure ($S_{m}$)~\cite{S-m} and E-measure ($E_{m}$)~\cite{Em} scores. The lower value is better for the MAE and the higher is better for others. Generally speaking, large-scale datasets can ensure the evaluation stability of model performance, while small-scale datasets often present large performance fluctuation. Ave-Metric has been used in HDFNet~\cite{HDFNet} to consider the scale of the data for comprehensive evaluation. 
\subsection{Implementation Details}
We first use random initialization to train the SSL cross-modal auto-encoder. And then the parameters of the encoder of each stream are loaded for the following depth-contour estimation. In this process, the parameters of the two-stream encoder are frozen, we only train the decoder with the CDA modules. Once the pretext tasks are finished, we load their parameters to initialize the network for the downstream task (RGB-D SOD). 

 Our models are implemented based on the Pytorch and trained on a  RTX 2080Ti GPU for $50$ epochs with mini-batch size $4$. We adopt some data augmentation techniques to avoid overfitting: random horizontally flipping, random rotate, random brightness, saturation and contrast. For the optimizer, we use the stochastic gradient descent (SGD) with a momentum of $0.9$ and a weight decay of $0.0005$. For the pretext tasks, the learning rate is set to $0.001$ and later use the ``poly'' policy~\cite{poly} with the power of $0.9$ as a means of adjustment. For the downstream task, maximum learning  rate  is  set  to $0.005$ for backbone and $0.05$ for other parts. Warm-up and linear decay strategies are used to adjust the learning rate.

\subsection{Comparisons with State-of-the-art}
For fair comparisons, the proposed algorithm is  compared with $16$ state-of-the-art methods that have released codes, including DisenFuse~\cite{DisenFuse}, A2DELE~\cite{A2DELE}, ICNet~\cite{ICNet}, S2MA~\cite{LSSA}, DANet~\cite{DANet}, CMWNet~\cite{CMWNet}, BBSNet~\cite{BBSNet}, CoNet~\cite{CoNet}, UCNet~\cite{UCNet}, PGAR~\cite{PGAR}, HDFNet~\cite{HDFNet}, CDNet~\cite{CDNet}, DFMNet~\cite{DFMNet}, DSA2F~\cite{DSA2F}, DCF~\cite{DCF} and HAINet~\cite{HAINet}. Saliency maps of these competitors are directly provided by their respective authors or computed by their released codes.

\textbf{Quantitative Evaluation.}
Tab.~\ref{tab:Table1} shows performance comparisons in terms of the F-measure, weighted F-measure, S-measure, E-measure and MAE scores.  According to the proportion of each testing set to all testing sets, the results on all datasets are weighted and summed to obtain an overall performance evaluation, which is listed in the row  ``Ave-Metric''. 

It can be seen that our fully-supervised (i.e., ImageNet pretrained) model ranks first among $18$ models in the overall ranking. Among $30$ scores of all datasets,  our $26$ scores reach the top-$3$ performance. It shows that our model has excellent comprehensive ability. Notably, our self-supervised model still can outperform most fully-supervised methods and ranks ninth among them, which shows the great potential of self-supervised learning and the effectiveness of the pretext tasks we design. 

\textbf{Qualitative Evaluation.}
 Fig.~\ref{fig:Figure6} illustrates the visual comparison with other RGB-D SOD approaches. The proposed method yields the results closer to the ground truth in various challenging scenarios. 
For the images with a single object, our method can completely segment the whole object, while other competitors lose more or less parts of the object (see the $1^{st}$ and $2^{nd}$ rows).
For images having multiple objects, our method can still accurately localize and capture all the objects (see the $3^{rd}$ and $4^{th}$ rows). Moreover, it can be seen that our SSL model even has better visual results compared to other fully supervised methods.




\subsection{Ablation Study}\label{sec:AbStd}
In this section, we detail the contribution of each component to the overall network. We first verify the effectiveness of the CDA module when the RGB-D SOD networks are pretrained on the ImageNet. Next, we evaluate the SSL pretext networks by loading different pretrained parameters to the RGB-D SOD network, the performance gain illustrates the benefits of the proposed SSL pretexts. All ablation experiments are based on the VGG-16 backbone.

\begin{table*}[ht]
\large

  \renewcommand\tabcolsep{5.0pt} 
  \renewcommand\arraystretch{1.5}
  \centering
\resizebox{\textwidth}{!}  
{
\begin{tabular}{cc|ccccccccccccccccc|c}
 \toprule[2pt]
\multicolumn{2}{l|}{\multirow{3}{*}{Metric}} & \multicolumn{17}{c|}{\textbf{{Fully-supervised (Pretrained on the ImageNet)}}}  & \multicolumn{1}{c}{\textbf{{Self-supervised}}} \\
\multicolumn{2}{l|}{}   & \Large{DisenFuse$_{20}$}         & \Large{A2DELE$_{20}$}         &\Large{ICNet$_{20}$} &\Large{S2MA$_{20}$} &\Large{DANet$_{20}$} & \Large{CMWNet$_{20}$} & \Large{BBSNet$_{20}$} &\Large{CoNet$_{20}$} & \Large{UCNet$_{20}$} & \Large{PGAR$_{20}$} &\Large{HDFNet$_{20}$} &\Large{CDNet$_{21}$} &\Large{DFMNet$_{21}$} &\Large{DSA2F$_{21}$} &\Large{DCF$_{21}$}&\Large{HAINet$_{21}$} &\Large{Ours} & \Large{Ours} 
\\
\multicolumn{2}{l|}{}   & VGG-16 & VGG-16 & VGG-16& VGG-16& VGG-16& VGG-16& VGG-16&  ResNet-101& VGG-16&  VGG-16& VGG-16& VGG-16 &MobileNet-V2& VGG-19& ResNet-18& VGG-16& VGG-16& VGG-16
\\
\hline
\multirow{5}{*}{\emph{\rotatebox{90}{RGBD135}}}      
&$F_{\beta}^{max}\uparrow$  &   \multicolumn{1}{c}{\Large{0.877}}    & \multicolumn{1}{c}{\Large{0.897}}  &\multicolumn{1}{c}{\Large{0.925}}  &  \multicolumn{1}{c}{\Large\textbf{0.944}}      &  \multicolumn{1}{c}{\Large{0.916}}      &  \multicolumn{1}{c}{\Large\textbf{0.939}}  &  \multicolumn{1}{c}{\Large{0.923}} &  \multicolumn{1}{c}{\Large{0.915}} &  \multicolumn{1}{c}{{\Large{0.936}}} &  \multicolumn{1}{c}{\Large{0.926}} &  \multicolumn{1}{c}{\Large{0.934}} &  \multicolumn{1}{c}{\Large\textbf{0.939}}&  \multicolumn{1}{c}{\Large{0.930}}&    \multicolumn{1}{c}{{\Large{0.930}}}  &  \multicolumn{1}{c}{{\Large{0.926}}} &  \multicolumn{1}{c}{{\Large{0.936}}}   &  \multicolumn{1}{c|}{\textbf{\Large{0.944}}}   &  \multicolumn{1}{c}{\textbf{\Large{0.941}}}     \\
&$F_{\beta}^{w}\uparrow$    &   \multicolumn{1}{c}{\Large{0.779}}    & \multicolumn{1}{c}{\Large{0.836}}  &\multicolumn{1}{c}{\Large{0.867}}  &  \multicolumn{1}{c}{\Large{0.892}}      &  \multicolumn{1}{c}{\Large{0.848}}   &  \multicolumn{1}{c}{\Large{0.888}}       &  \multicolumn{1}{c}{\Large{0.845}}    &  \multicolumn{1}{c}{\Large{0.849}}  &  \multicolumn{1}{c}{\textbf{\Large{0.908}}}  &    \multicolumn{1}{c}{\Large{0.855}} &  \multicolumn{1}{c}{\Large{0.902}}   &  \multicolumn{1}{c}{\Large{0.894}} &  \multicolumn{1}{c}{\Large{0.888}} &  \multicolumn{1}{c}{{\Large{0.882}}}     &   \multicolumn{1}{c}{{\Large{0.876}}}  &   \multicolumn{1}{c}{{\Large{0.898}}}      &   \multicolumn{1}{c|}{\textbf{\Large{0.917}}}      &   \multicolumn{1}{c}{\textbf{\Large{0.910}}}           \\
& $S_m\uparrow$          &   \multicolumn{1}{c}{\Large{0.779}}    & \multicolumn{1}{c}{\Large{0.886}} &\multicolumn{1}{c}{\Large{0.920}}  &  \multicolumn{1}{c}{\textbf{\Large{0.941}}}      &  \multicolumn{1}{c}{\Large{0.905}}   &  \multicolumn{1}{c}{\textbf{\Large{0.934}}} &  \multicolumn{1}{c}{\Large{0.908}} &  \multicolumn{1}{c}{\Large{0.909}}&  \multicolumn{1}{c}{\textbf{\Large{0.934}}} &  \multicolumn{1}{c}{\Large{0.913}}               &    \multicolumn{1}{c}{\Large{0.932}}&    \multicolumn{1}{c}{\textbf{\Large{0.936}}}&    \multicolumn{1}{c}{\Large{0.931}}&  \multicolumn{1}{c}{{\Large{0.916}}}         &   \multicolumn{1}{c}{{\Large{0.916}}}  &   \multicolumn{1}{c}{{\Large{0.929}}}  &   \multicolumn{1}{c|}{\textbf{\Large{0.936}}}  &   \multicolumn{1}{c}{{\Large{0.932}}}        \\
& $E_m\uparrow$      &   \multicolumn{1}{c}{\Large{0.923}}    & \multicolumn{1}{c}{\Large{0.920}}  &\multicolumn{1}{c}{\Large{0.959}}  &  \multicolumn{1}{c}{\textbf{\Large{0.974}}}      &  \multicolumn{1}{c}{\Large{0.961}}   &  \multicolumn{1}{c}{\Large{0.967}}      &  \multicolumn{1}{c}{\Large{0.941}}  &  \multicolumn{1}{c}{\Large{0.945}}     &  \multicolumn{1}{c}{\textbf{\Large{0.974}}}    &  \multicolumn{1}{c}{\Large{0.939}}  &  \multicolumn{1}{c}{{\Large{0.973}}}  &  \multicolumn{1}{c}{\Large{0.969}}&  \multicolumn{1}{c}{\Large{0.971}} &    \multicolumn{1}{c}{{\Large{0.955}}}    &   \multicolumn{1}{c}{{\Large{0.958}}}    &   \multicolumn{1}{c}{{\Large{0.967}}} &   \multicolumn{1}{c|}{\textbf{\Large{0.978}}} &   \multicolumn{1}{c}{\textbf{\Large{0.976}}}    \\
&$\mathcal{M}\downarrow$  &   \multicolumn{1}{c}{\Large{0.040}}    & \multicolumn{1}{c}{\Large{0.029}}  &\multicolumn{1}{c}{\Large{0.027}}  &  \multicolumn{1}{c}{\Large{0.021}}      &  \multicolumn{1}{c}{\Large{0.028}}    &  \multicolumn{1}{c}{\Large{0.022}}    &  \multicolumn{1}{c}{\Large{0.029}}   &  \multicolumn{1}{c}{\Large{0.028}}  &  \multicolumn{1}{c}{\textbf{\Large{0.019}}} &  \multicolumn{1}{c}{\Large{0.026}}           &    \multicolumn{1}{c}{\textbf{\Large{0.020}}}   &    \multicolumn{1}{c}{\textbf{\Large{0.020}}} &    \multicolumn{1}{c}{\Large{0.021}}  &   \multicolumn{1}{c}{{\Large{0.023}}}   &  \multicolumn{1}{c}{{\Large{0.023}}}  &  \multicolumn{1}{c}{\textbf{\Large{0.019}}}&  \multicolumn{1}{c|}{\textbf{\Large{0.017}}}&  \multicolumn{1}{c}{\textbf{\Large{0.019}}}    \\
\hline
\multirow{5}{*}{\emph{\rotatebox{90}{DUT-RGBD}}}      
&$F_{\beta}^{max}\uparrow$   &   \multicolumn{1}{c}{\Large{0.807}}    & \multicolumn{1}{c}{\Large{0.907}}  &\multicolumn{1}{c}{\Large{0.875}}  &  \multicolumn{1}{c}{\Large{0.909}}      &  \multicolumn{1}{c}{\Large{0.911}}   &  \multicolumn{1}{c}{\Large{0.905}} &  \multicolumn{1}{c}{\Large{-}}&  \multicolumn{1}{c}{\Large{0.935}} &  \multicolumn{1}{c}{\Large{-}}&  \multicolumn{1}{c}{\textbf{\Large{0.938}}}     &    \multicolumn{1}{c}{\Large{0.926}}&    \multicolumn{1}{c}{\Large{0.901}}&    \multicolumn{1}{c}{\Large{-}}       &   \multicolumn{1}{c}{\textbf{\Large{0.938}}} &  \multicolumn{1}{c}{\textbf{\Large{0.941}}}  &  \multicolumn{1}{c}{{\Large{0.932}}} &  \multicolumn{1}{c|}{\textbf{\Large{0.947}}} &  \multicolumn{1}{c}{{\Large{0.930}}} \\
&$F_{\beta}^{w}\uparrow$    &   \multicolumn{1}{c}{\Large{0.691}}    & \multicolumn{1}{c}{\Large{0.864}}  &\multicolumn{1}{c}{\Large{0.784}}  &  \multicolumn{1}{c}{\Large{0.862}}      &  \multicolumn{1}{c}{\Large{0.847}}    &  \multicolumn{1}{c}{\Large{0.831}}  &  \multicolumn{1}{c}{\Large{-}} &  \multicolumn{1}{c}{{\Large{0.891}}} &  \multicolumn{1}{c}{\Large{-}}&  \multicolumn{1}{c}{\Large{0.889}}    &  \multicolumn{1}{c}{\Large{0.865}}  &  \multicolumn{1}{c}{\Large{0.838}}  &  \multicolumn{1}{c}{\Large{-}}     &    \multicolumn{1}{c}{\textbf{\Large{0.908}}}      &   \multicolumn{1}{c}{\textbf{\Large{0.909}}}   &   \multicolumn{1}{c}{{\Large{0.883}}}&   \multicolumn{1}{c|}{\textbf{\Large{0.914}}}&   \multicolumn{1}{c}{{\Large{0.883}}} \\
& $S_m\uparrow$  &   \multicolumn{1}{c}{\Large{0.798}}    & \multicolumn{1}{c}{\Large{0.886}} &\multicolumn{1}{c}{\Large{0.852}}  &  \multicolumn{1}{c}{\Large{0.903}}      &  \multicolumn{1}{c}{\Large{0.889}}   &  \multicolumn{1}{c}{\Large{0.887}} &  \multicolumn{1}{c}{\Large{-}} &  \multicolumn{1}{c}{\Large{0.919}} &  \multicolumn{1}{c}{{\Large{-}}} &  \multicolumn{1}{c}{{\Large{0.920}}} &  \multicolumn{1}{c}{\Large{0.905}}  &  \multicolumn{1}{c}{\Large{0.886}}&  \multicolumn{1}{c}{\Large{-}} &    \multicolumn{1}{c}{\textbf{\Large{0.921}}}       &   \multicolumn{1}{c}{\textbf{\Large{0.924}}}   &   \multicolumn{1}{c}{{\Large{0.909}}} &   \multicolumn{1}{c|}{\textbf{\Large{0.929}}} &   \multicolumn{1}{c}{{\Large{0.908}}}   \\
& $E_m\uparrow$    &   \multicolumn{1}{c}{\Large{0.854}}    & \multicolumn{1}{c}{\Large{0.929}}  &\multicolumn{1}{c}{\Large{0.901}}  &  \multicolumn{1}{c}{\Large{0.921}}      &  \multicolumn{1}{c}{\Large{0.929}}    &  \multicolumn{1}{c}{{\Large{0.922}}}   &  \multicolumn{1}{c}{\Large{-}}  &  \multicolumn{1}{c}{\Large{0.952}}  &  \multicolumn{1}{c}{\Large{-}}  &  \multicolumn{1}{c}{\Large{0.950}}  &  \multicolumn{1}{c}{\Large{0.938}}&  \multicolumn{1}{c}{\Large{0.917}}&  \multicolumn{1}{c}{\Large{-}}&    \multicolumn{1}{c}{\textbf{\Large{0.956}}}       &   \multicolumn{1}{c}{\textbf{\Large{0.957}}}      &   \multicolumn{1}{c}{{\Large{0.939}}}   &   \multicolumn{1}{c|}{\textbf{\Large{0.958}}}   &   \multicolumn{1}{c}{{\Large{0.943}}}    \\
&$\mathcal{M}\downarrow$  &   \multicolumn{1}{c}{\Large{0.104}}    & \multicolumn{1}{c}{\Large{0.043}}  &\multicolumn{1}{c}{\Large{0.072}}  &  \multicolumn{1}{c}{\Large{0.044}}      &  \multicolumn{1}{c}{\Large{0.047}}   &  \multicolumn{1}{c}{\Large{0.056}}    &  \multicolumn{1}{c}{\Large{-}}  &  \multicolumn{1}{c}{{\Large{0.033}}}  &  \multicolumn{1}{c}{\Large{-}} &  \multicolumn{1}{c}{\Large{0.035}}  &  \multicolumn{1}{c}{\Large{0.040}} &  \multicolumn{1}{c}{\Large{0.051}} &  \multicolumn{1}{c}{\Large{-}}   &    \multicolumn{1}{c}{\textbf{\Large{0.031}}}      &   \multicolumn{1}{c}{\textbf{\Large{0.030}}}    &   \multicolumn{1}{c}{{\Large{0.038}}}   &   \multicolumn{1}{c|}{\textbf{\Large{0.029}}}   &   \multicolumn{1}{c}{{\Large{0.038}}}     \\

\hline
\multirow{5}{*}{\emph{\rotatebox{90}{STERE}}}      
&$F_{\beta}^{max}\uparrow$   &   \multicolumn{1}{c}{\Large{0.887}}    & \multicolumn{1}{c}{\Large{0.892}}  &\multicolumn{1}{c}{\Large{0.897}}  &  \multicolumn{1}{c}{\Large{0.895}}      &  \multicolumn{1}{c}{\Large{0.897}}   &  \multicolumn{1}{c}{\Large{0.911}} &  \multicolumn{1}{c}{{\Large{0.901}}}&  \multicolumn{1}{c}{\Large{0.909}} &  \multicolumn{1}{c}{\Large{0.908}}&  \multicolumn{1}{c}{\Large{0.911}}     &    \multicolumn{1}{c}{\textbf{\Large{0.918}}}&    \multicolumn{1}{c}{{\Large{0.909}}}&    \multicolumn{1}{c}{{\Large{0.902}}}       &   \multicolumn{1}{c}{{\Large{0.910}}} &  \multicolumn{1}{c}{\textbf{\Large{0.915}}}&  \multicolumn{1}{c}{\textbf{\Large{0.919}}}&  \multicolumn{1}{c|}{\Large{0.914}}&  \multicolumn{1}{c}{\Large{0.897}}  \\
&$F_{\beta}^{w}\uparrow$    &   \multicolumn{1}{c}{\Large{0.811}}    & \multicolumn{1}{c}{\Large{0.846}}  &\multicolumn{1}{c}{\Large{0.815}}  &  \multicolumn{1}{c}{\Large{0.825}}      &  \multicolumn{1}{c}{\Large{0.830}}    &  \multicolumn{1}{c}{\Large{0.847}}  &  \multicolumn{1}{c}{\Large{0.838}} &  \multicolumn{1}{c}{{\Large{0.866}}} &  \multicolumn{1}{c}{\Large{0.867}}&  \multicolumn{1}{c}{\Large{0.856}}    &  \multicolumn{1}{c}{{\Large{0.863}}}  &  \multicolumn{1}{c}{{\Large{0.855}}}  &  \multicolumn{1}{c}{{\Large{0.841}}}     &    \multicolumn{1}{c}{{\Large{0.869}}}      &   \multicolumn{1}{c}{\textbf{\Large{0.873}}}    &   \multicolumn{1}{c}{\textbf{\Large{0.871}}}  &   \multicolumn{1}{c|}{\textbf{\Large{0.870}}}  &   \multicolumn{1}{c}{\Large{0.845}}  \\
& $S_m\uparrow$  &   \multicolumn{1}{c}{\Large{0.883}}    & \multicolumn{1}{c}{\Large{0.878}} &\multicolumn{1}{c}{\Large{0.891}}  &  \multicolumn{1}{c}{\Large{0.890}}      &  \multicolumn{1}{c}{\Large{0.892}}   &  \multicolumn{1}{c}{{\Large{0.905}}} &  \multicolumn{1}{c}{{\Large{0.896}}} &  \multicolumn{1}{c}{{\Large{0.905}}} &  \multicolumn{1}{c}{\Large{0.903}} &  \multicolumn{1}{c}{\textbf{\Large{0.907}}} &  \multicolumn{1}{c}{\textbf{\Large{0.906}}}  &  \multicolumn{1}{c}{{\Large{0.903}}} &  \multicolumn{1}{c}{{\Large{0.898}}}  &    \multicolumn{1}{c}{{\Large{0.897}}}       &   \multicolumn{1}{c}{{\Large{0.905}}}    &   \multicolumn{1}{c}{\textbf{\Large{0.909}}}&   \multicolumn{1}{c|}{\Large{0.904}}&   \multicolumn{1}{c}{\Large{0.885}} \\
& $E_m\uparrow$    &   \multicolumn{1}{c}{\Large{0.915}}    & \multicolumn{1}{c}{\Large{0.928}}  &\multicolumn{1}{c}{\Large{0.911}}  &  \multicolumn{1}{c}{\Large{0.926}}      &  \multicolumn{1}{c}{\Large{0.927}}    &  \multicolumn{1}{c}{\Large{0.930}}   &  \multicolumn{1}{c}{\Large{0.928}}  &  \multicolumn{1}{c}{\textbf{\Large{0.941}}}  &  \multicolumn{1}{c}{\textbf{\Large{0.942}}}  &  \multicolumn{1}{c}{\Large{0.937}}  &  \multicolumn{1}{c}{{\Large{0.937}}}&  \multicolumn{1}{c}{\Large{0.938}}&  \multicolumn{1}{c}{\Large{0.931}}&    \multicolumn{1}{c}{\textbf{\Large{0.942}}}       &   \multicolumn{1}{c}{\textbf{\Large{0.943}}}       &   \multicolumn{1}{c}{{\Large{0.938}}}  &   \multicolumn{1}{c|}{{\Large{0.939}}}  &   \multicolumn{1}{c}{{\Large{0.929}}}  \\
&$\mathcal{M}\downarrow$  &   \multicolumn{1}{c}{\Large{0.054}}    & \multicolumn{1}{c}{\Large{0.045}}  &\multicolumn{1}{c}{\Large{0.054}}  &  \multicolumn{1}{c}{\Large{0.051}}      &  \multicolumn{1}{c}{\Large{0.048}}   &  \multicolumn{1}{c}{\Large{0.043}}    &  \multicolumn{1}{c}{{\Large{0.046}}}  &  \multicolumn{1}{c}{\textbf{\Large{0.037}}}  &  \multicolumn{1}{c}{\textbf{\Large{0.039}}} &  \multicolumn{1}{c}{\Large{0.041}}  &  \multicolumn{1}{c}{\textbf{\Large{0.039}}} &  \multicolumn{1}{c}{\Large{0.041}} &  \multicolumn{1}{c}{\Large{0.045}}   &    \multicolumn{1}{c}{\textbf{\Large{0.039}}}      &   \multicolumn{1}{c}{\textbf{\Large{0.037}}}      &   \multicolumn{1}{c}{\textbf{\Large{0.038}}}    &   \multicolumn{1}{c|}{\textbf{\Large{0.039}}}    &   \multicolumn{1}{c}{\Large{0.047}}    \\
\hline
\multirow{5}{*}{\emph{\rotatebox{90}{NLPR}}}      
&$F_{\beta}^{max}\uparrow$    &   \multicolumn{1}{c}{\Large{0.895}}    & \multicolumn{1}{c}{\Large{0.898}}  &\multicolumn{1}{c}{\Large{0.919}}  &  \multicolumn{1}{c}{\Large{0.910}}      &  \multicolumn{1}{c}{\Large{0.908}}   &  \multicolumn{1}{c}{\Large{0.913}} &  \multicolumn{1}{c}{{\textbf{\Large{0.921}}}}  &  \multicolumn{1}{c}{\Large{0.898}} &  \multicolumn{1}{c}{\Large{0.916}} &  \multicolumn{1}{c}{\textbf{\Large{0.925}}}  &  \multicolumn{1}{c}{\Large{0.917}} &  \multicolumn{1}{c}{\textbf{\Large{0.925}}}&  \multicolumn{1}{c}{\Large{0.916}}&    \multicolumn{1}{c}{{\Large{0.916}}}     &   \multicolumn{1}{c}{{\Large{0.917}}}     &   \multicolumn{1}{c}{{\Large{0.917}}}   &   \multicolumn{1}{c|}{\textbf{\Large{0.923}}}   &   \multicolumn{1}{c}{{\Large{0.912}}}    \\
&$F_{\beta}^{w}\uparrow$    &   \multicolumn{1}{c}{\Large{0.828}}    & \multicolumn{1}{c}{\Large{0.857}}  &\multicolumn{1}{c}{\Large{0.864}}  &  \multicolumn{1}{c}{\Large{0.852}}      &  \multicolumn{1}{c}{\Large{0.850}}   &  \multicolumn{1}{c}{\Large{0.856}}&  \multicolumn{1}{c}{{\Large{0.871}}}&  \multicolumn{1}{c}{\Large{0.842}} &  \multicolumn{1}{c}{\Large{0.878}}    &  \multicolumn{1}{c}{\Large{0.881}}    &  \multicolumn{1}{c}{\Large{0.869}}   &  \multicolumn{1}{c}{\textbf{\Large{0.882}}} &  \multicolumn{1}{c}{\Large{0.869}}               &    \multicolumn{1}{c}{{\Large{0.881}}}    &   \multicolumn{1}{c}{\textbf{\Large{0.886}}}     &   \multicolumn{1}{c}{{\Large{0.881}}}  &   \multicolumn{1}{c|}{\textbf{\Large{0.889}}}  &   \multicolumn{1}{c}{{\Large{0.871}}}    \\
& $S_m\uparrow  $         &   \multicolumn{1}{c}{\Large{0.900}}    & \multicolumn{1}{c}{\Large{0.898}}  &\multicolumn{1}{c}{\textbf{\Large{0.922}}}  &  \multicolumn{1}{c}{\Large{0.915}}      &  \multicolumn{1}{c}{\Large{0.908}}    &  \multicolumn{1}{c}{\Large{0.917}}  &  \multicolumn{1}{c}{\textbf{\Large{0.923}}} &  \multicolumn{1}{c}{\Large{0.908}} &  \multicolumn{1}{c}{\Large{0.920}}  &  \multicolumn{1}{c}{\textbf{\Large{0.930}}} &  \multicolumn{1}{c}{{\Large{0.916}}} &  \multicolumn{1}{c}{\textbf{\Large{0.930}}} &  \multicolumn{1}{c}{\textbf{\Large{0.923}}}&    \multicolumn{1}{c}{{\Large{0.918}}}       &   \multicolumn{1}{c}{\Large{0.921}}      &   \multicolumn{1}{c}{\Large{0.921}}    &   \multicolumn{1}{c|}{\textbf{\Large{0.922}}}    &   \multicolumn{1}{c}{\Large{0.913}}    \\
& $E_m\uparrow$  &   \multicolumn{1}{c}{\Large{0.933}}    & \multicolumn{1}{c}{\Large{0.945}}  &\multicolumn{1}{c}{\Large{0.945}}  &  \multicolumn{1}{c}{\Large{0.942}}      &  \multicolumn{1}{c}{\Large{0.945}}   &  \multicolumn{1}{c}{\Large{0.941}}    &  \multicolumn{1}{c}{{\Large{0.948}}}     &  \multicolumn{1}{c}{{\Large{0.934}}} &  \multicolumn{1}{c}{\textbf{\Large{0.955}}}    &  \multicolumn{1}{c}{\textbf{\Large{0.955}}}  &  \multicolumn{1}{c}{\Large{0.948}} &  \multicolumn{1}{c}{\Large{0.954}}&  \multicolumn{1}{c}{\Large{0.951}}         &    \multicolumn{1}{c}{{\Large{0.952}}}    &   \multicolumn{1}{c}{\textbf{\Large{0.956}}}   &   \multicolumn{1}{c}{{\Large{0.952}}}  &   \multicolumn{1}{c|}{\textbf{\textbf{\Large{0.960}}}}  &   \multicolumn{1}{c}{{\Large{0.947}}}      \\
&$\mathcal{M}\downarrow$    &   \multicolumn{1}{c}{\Large{0.035}}    & \multicolumn{1}{c}{\Large{0.029}}  &\multicolumn{1}{c}{\Large{0.028}}  &  \multicolumn{1}{c}{\Large{0.030}}      &  \multicolumn{1}{c}{\Large{0.031}}    &  \multicolumn{1}{c}{\Large{0.029}}  &  \multicolumn{1}{c}{{\Large{0.026}}}  &  \multicolumn{1}{c}{{\Large{0.031}}}  &  \multicolumn{1}{c}{\textbf{\Large{0.025}}}  &  \multicolumn{1}{c}{\textbf{\Large{0.024}}}  &  \multicolumn{1}{c}{{\Large{0.027}}}&  \multicolumn{1}{c}{\textbf{\Large{0.024}}} &  \multicolumn{1}{c}{\Large{0.026}} &    \multicolumn{1}{c}{\textbf{\Large{0.024}}}       &   \multicolumn{1}{c}{\textbf{\Large{0.023}}}     &   \multicolumn{1}{c}{\textbf{\Large{0.025}}}   &   \multicolumn{1}{c|}{\textbf{\Large{0.025}}}   &   \multicolumn{1}{c}{{\Large{0.027}}}    \\
\hline
\multirow{5}{*}{\emph{\rotatebox{90}{NJUD}}}      
&$F_{\beta}^{max}\uparrow$   &   \multicolumn{1}{c}{\Large{0.897}}    & \multicolumn{1}{c}{\Large{0.890}}  &\multicolumn{1}{c}{\Large{0.903}}  &  \multicolumn{1}{c}{\Large{0.899}}      &  \multicolumn{1}{c}{\Large{0.905}}    &  \multicolumn{1}{c}{\Large{0.913}}  &  \multicolumn{1}{c}{\textbf{\Large{0.926}}}  &  \multicolumn{1}{c}{\Large{0.902}}  &  \multicolumn{1}{c}{\Large{0.908}} &  \multicolumn{1}{c}{\Large{0.918}}     &  \multicolumn{1}{c}{\textbf{\Large{0.924}}}&  \multicolumn{1}{c}{\Large{0.919}}  &  \multicolumn{1}{c}{\Large{0.918}}   &    \multicolumn{1}{c}{{\Large{0.917}}}      &   \multicolumn{1}{c}{{\Large{0.917}}}   &   \multicolumn{1}{c}{\Large{0.920}} &   \multicolumn{1}{c|}{\textbf{\Large{0.923}}} &   \multicolumn{1}{c}{{\Large{0.913}}}  \\
&$F_{\beta}^{w}\uparrow$     &   \multicolumn{1}{c}{\Large{0.831}}    & \multicolumn{1}{c}{\Large{0.843}}  &\multicolumn{1}{c}{\Large{0.833}}  &  \multicolumn{1}{c}{\Large{0.842}}      &  \multicolumn{1}{c}{\Large{0.853}}     &  \multicolumn{1}{c}{\Large{0.857}}   &  \multicolumn{1}{c}{\Large{0.878}}   &  \multicolumn{1}{c}{\Large{0.850}}      &  \multicolumn{1}{c}{\Large{0.868}}   &  \multicolumn{1}{c}{\Large{0.872}}   &  \multicolumn{1}{c}{\textbf{\Large{0.881}}}    &  \multicolumn{1}{c}{\Large{0.878}}  &  \multicolumn{1}{c}{\Large{0.868}}    &    \multicolumn{1}{c}{\textbf{\Large{0.883}}}    &   \multicolumn{1}{c}{{\Large{0.878}}}    &   \multicolumn{1}{c}{{\Large{0.879}}} &   \multicolumn{1}{c|}{\textbf{\Large{0.884}}} &   \multicolumn{1}{c}{{\Large{0.872}}}    \\
& $S_m\uparrow$        &   \multicolumn{1}{c}{\Large{0.889}}    & \multicolumn{1}{c}{\Large{0.871}} &\multicolumn{1}{c}{\Large{0.895}}  &  \multicolumn{1}{c}{\Large{0.894}}      &  \multicolumn{1}{c}{\Large{0.897}}    &  \multicolumn{1}{c}{\Large{0.903}}  &  \multicolumn{1}{c}{\textbf{\Large{0.916}}}  &  \multicolumn{1}{c}{\Large{0.895}}  &  \multicolumn{1}{c}{\Large{0.897}} &  \multicolumn{1}{c}{\Large{0.909}}  &  \multicolumn{1}{c}{\textbf{\Large{0.911}}}  &  \multicolumn{1}{c}{\textbf{\Large{0.913}}}&  \multicolumn{1}{c}{\Large{0.906}}  &    \multicolumn{1}{c}{{\Large{0.904}}}      &   \multicolumn{1}{c}{{\Large{0.903}}}    &   \multicolumn{1}{c}{{\Large{0.909}}}  &   \multicolumn{1}{c|}{{\Large{0.909}}}  &   \multicolumn{1}{c}{{\Large{0.901}}}     \\
& $E_m\uparrow$     &   \multicolumn{1}{c}{\Large{0.914}}    & \multicolumn{1}{c}{\Large{0.916}}  &\multicolumn{1}{c}{\Large{0.901}}  &  \multicolumn{1}{c}{\Large{0.917}}      &  \multicolumn{1}{c}{\Large{0.926}}    &  \multicolumn{1}{c}{\Large{0.923}}     &  \multicolumn{1}{c}{\Large{0.937}}  &  \multicolumn{1}{c}{\Large{0.924}}  &  \multicolumn{1}{c}{\Large{0.934}}  &  \multicolumn{1}{c}{{\Large{0.935}}}  &  \multicolumn{1}{c}{{\Large{0.934}}}   &  \multicolumn{1}{c}{\textbf{\Large{0.940}}}&  \multicolumn{1}{c}{\Large{0.934}}   &    \multicolumn{1}{c}{\Large{0.937}}     &   \multicolumn{1}{c}{\textbf{\Large{0.941}}}    &   \multicolumn{1}{c}{{\Large{0.931}}}  &   \multicolumn{1}{c|}{\textbf{\Large{0.939}}}  &   \multicolumn{1}{c}{{\Large{0.932}}}    \\
&$\mathcal{M}\downarrow$  &   \multicolumn{1}{c}{\Large{0.052}}    & \multicolumn{1}{c}{\Large{0.047}}  &\multicolumn{1}{c}{\Large{0.051}}  &  \multicolumn{1}{c}{\Large{0.053}}      &  \multicolumn{1}{c}{\Large{0.046}}   &  \multicolumn{1}{c}{\Large{0.046}}   &  \multicolumn{1}{c}{\textbf{\Large{0.039}}}   &  \multicolumn{1}{c}{\Large{0.046}}  &  \multicolumn{1}{c}{\Large{0.043}}  &  \multicolumn{1}{c}{{\Large{0.042}}}  &  \multicolumn{1}{c}{\textbf{\Large{0.037}}} &  \multicolumn{1}{c}{\textbf{\Large{0.038}}} &  \multicolumn{1}{c}{\Large{0.042}}  &    \multicolumn{1}{c}{\textbf{\Large{0.039}}}      &   \multicolumn{1}{c}{\textbf{\Large{0.038}}}   &   \multicolumn{1}{c}{\textbf{\Large{0.038}}} &   \multicolumn{1}{c|}{\textbf{\Large{0.038}}} &   \multicolumn{1}{c}{{\Large{0.042}}}    \\
\hline
\multirow{5}{*}{\emph{\rotatebox{90}{SIP}}}      
&$F_{\beta}^{max}\uparrow$   &   \multicolumn{1}{c}{{\Large{0.866}}}    & \multicolumn{1}{c}{\Large{0.855}}  &\multicolumn{1}{c}{\Large{0.882}}  &  \multicolumn{1}{c}{\Large{0.891}}      &  \multicolumn{1}{c}{\Large{0.901}}    &  \multicolumn{1}{c}{\Large{0.890}}  &  \multicolumn{1}{c}{\Large{0.892}} &  \multicolumn{1}{c}{\Large{0.883}}    &  \multicolumn{1}{c}{{\Large{0.896}}} &  \multicolumn{1}{c}{{\Large{0.893}}} &  \multicolumn{1}{c}{\textbf{\Large{0.904}}}  &  \multicolumn{1}{c}{\Large{0.888}} &  \multicolumn{1}{c}{\textbf{\Large{0.904}}}   &    \multicolumn{1}{c}{{\Large{0.891}}}      &   \multicolumn{1}{c}{\Large{0.900}}    &   \multicolumn{1}{c}{\textbf{{\Large{0.916}}}}  &   \multicolumn{1}{c|}{\textbf{\Large{0.909}}}  &   \multicolumn{1}{c}{\Large{0.895}}   \\
&$F_{\beta}^{w}\uparrow$   &   \multicolumn{1}{c}{\Large{0.780}}    & \multicolumn{1}{c}{\Large{0.780}}  &\multicolumn{1}{c}{\Large{0.793}}  &  \multicolumn{1}{c}{\Large{0.819}}      &  \multicolumn{1}{c}{\Large{0.829}}    &  \multicolumn{1}{c}{\Large{0.811}}  &  \multicolumn{1}{c}{\Large{0.820}}  &  \multicolumn{1}{c}{\Large{0.803}}  &  \multicolumn{1}{c}{{\Large{0.836}}} &  \multicolumn{1}{c}{{\Large{0.822}}}  &  \multicolumn{1}{c}{\Large{0.835}}  &  \multicolumn{1}{c}{\Large{0.812}}&  \multicolumn{1}{c}{\Large{0.835}}  &    \multicolumn{1}{c}{{\Large{0.829}}}      &   \multicolumn{1}{c}{\textbf{\Large{0.841}}}     &   \multicolumn{1}{c}{\textbf{\Large{0.854}}}  &   \multicolumn{1}{c|}{\textbf{\Large{0.855}}}  &   \multicolumn{1}{c}{\Large{0.830}}   \\
& $S_m\uparrow$        &   \multicolumn{1}{c}{\Large{0.859}}    & \multicolumn{1}{c}{\Large{0.828}} &\multicolumn{1}{c}{\Large{0.864}}  &  \multicolumn{1}{c}{\Large{0.872}}      &  \multicolumn{1}{c}{\Large{0.878}}   &  \multicolumn{1}{c}{\Large{0.867}}   &  \multicolumn{1}{c}{\Large{0.874}}    &  \multicolumn{1}{c}{\Large{0.858}}    &  \multicolumn{1}{c}{{\Large{0.875}}}&  \multicolumn{1}{c}{{\Large{0.876}}}    &  \multicolumn{1}{c}{\Large{0.878}} &  \multicolumn{1}{c}{\Large{0.862}}&  \multicolumn{1}{c}{\textbf{\Large{0.883}}}    &    \multicolumn{1}{c}{{\Large{0.862}}}     &   \multicolumn{1}{c}{\Large{0.873}}    &   \multicolumn{1}{c}{\textbf{\Large{0.886}}}&   \multicolumn{1}{c|}{\textbf{\Large{0.888}}}&   \multicolumn{1}{c}{\Large{0.868}}   \\
& $E_m\uparrow$     &   \multicolumn{1}{c}{\Large{0.899}}    & \multicolumn{1}{c}{\Large{0.890}}  &\multicolumn{1}{c}{\Large{0.903}}  &  \multicolumn{1}{c}{\Large{0.913}}      &  \multicolumn{1}{c}{\Large{0.917}}    &  \multicolumn{1}{c}{\Large{0.909}}    &  \multicolumn{1}{c}{\Large{0.912}}     &  \multicolumn{1}{c}{\Large{0.909}}  &  \multicolumn{1}{c}{{\Large{0.918}}}  &  \multicolumn{1}{c}{{\Large{0.912}}}  &  \multicolumn{1}{c}{\Large{0.921}}  &  \multicolumn{1}{c}{\Large{0.905}}&  \multicolumn{1}{c}{\textbf{\Large{0.923}}}        &    \multicolumn{1}{c}{\Large{0.911}}    &   \multicolumn{1}{c}{\Large{0.921}} &   \multicolumn{1}{c}{\textbf{\Large{0.925}}} &   \multicolumn{1}{c|}{\textbf{\Large{0.927}}} &   \multicolumn{1}{c}{\Large{0.910}}       \\
&$\mathcal{M}\downarrow$  &   \multicolumn{1}{c}{\Large{0.068}}    & \multicolumn{1}{c}{\Large{0.070}}  &\multicolumn{1}{c}{\Large{0.063}}  &  \multicolumn{1}{c}{\Large{0.057}}      &  \multicolumn{1}{c}{\Large{0.054}}  &  \multicolumn{1}{c}{\Large{0.062}}    &  \multicolumn{1}{c}{\Large{0.056}}    &  \multicolumn{1}{c}{\Large{0.063}}    &  \multicolumn{1}{c}{{\Large{0.051}}} &  \multicolumn{1}{c}{{\Large{0.055}}}  &  \multicolumn{1}{c}{\textbf{\Large{0.050}}} &  \multicolumn{1}{c}{\Large{0.060}} &  \multicolumn{1}{c}{\Large{0.051}}      &    \multicolumn{1}{c}{{\Large{0.057}}}     &   \multicolumn{1}{c}{\Large{0.052}}   &   \multicolumn{1}{c}{\textbf{\Large{0.048}}}&   \multicolumn{1}{c|}{\textbf{\Large{0.046}}}&   \multicolumn{1}{c}{\Large{0.058}}   \\

\bottomrule[1pt]
\bottomrule[2pt]
\rowcolor{mygray}
&$Top$ $3$   &   
\multicolumn{1}{c}{\Large{0/30}}    &
\multicolumn{1}{c}{{\Large{0/30}}}    &
\multicolumn{1}{c}{{\Large{1/30}}}    &
\multicolumn{1}{c}{{\Large{3/30}}}    &
\multicolumn{1}{c}{{\Large{0/30}}}    &
\multicolumn{1}{c}{{\Large{2/30}}}    &
\multicolumn{1}{c}{{\Large{5/25}}}    &
\multicolumn{1}{c}{{\Large{2/30}}}    &
\multicolumn{1}{c}{{\Large{8/25}}}    &
\multicolumn{1}{c}{{\Large{6/30}}}    &
\multicolumn{1}{c}{{\Large{10/30}}}    &
\multicolumn{1}{c}{{\Large{10/30}}}    &
\multicolumn{1}{c}{{\Large{4/25}}}    &
\multicolumn{1}{c}{{\Large{10/30}}}    &
\multicolumn{1}{c}{{\Large{15/30}}}    &
\multicolumn{1}{c}{{\Large{12/30}}}    &
\multicolumn{1}{c|}{{\Large{26/30}}}    &
\multicolumn{1}{c}{{\Large{4/30}}}    
\\
\bottomrule[1pt]
\bottomrule[2pt]
\multirow{5}{*}{\emph{\rotatebox{90}{Ave-Metric}}}      
&$F_{\beta}^{max}\uparrow$   &   \multicolumn{1}{c}{\Large{0.873}}    & \multicolumn{1}{c}{\Large{0.884}}  &\multicolumn{1}{c}{\Large{0.894}}  &  \multicolumn{1}{c}{\Large{0.900}}      &  \multicolumn{1}{c}{\Large{0.903}}    &  \multicolumn{1}{c}{\Large{0.906}}  &  \multicolumn{1}{c}{\Large{0.906}} &  \multicolumn{1}{c}{\Large{0.903}}    &  \multicolumn{1}{c}{\Large{0.906}} &  \multicolumn{1}{c}{\Large{0.912}} &  \multicolumn{1}{c}{{\Large{0.916}}}    &    \multicolumn{1}{c}{{\Large{0.906}}}      &   \multicolumn{1}{c}{{\Large{0.908}}}   &   \multicolumn{1}{c}{{\Large{0.910}}}  &   \multicolumn{1}{c}{{\Large{0.915}}}  &   \multicolumn{1}{c}{{\Large{0.920}}}  &   \multicolumn{1}{c|}{{\Large{0.920}}}  &   \multicolumn{1}{c}{{\Large{0.906}}}    \\
&$F_{\beta}^{w}\uparrow$   &   \multicolumn{1}{c}{\Large{0.791}}    & \multicolumn{1}{c}{\Large{0.830}}  &\multicolumn{1}{c}{\Large{0.815}}  &  \multicolumn{1}{c}{\Large{0.835}}      &  \multicolumn{1}{c}{\Large{0.838}}    &  \multicolumn{1}{c}{\Large{0.839}}  &  \multicolumn{1}{c}{\Large{0.843}}  &  \multicolumn{1}{c}{\Large{0.846}}  &  \multicolumn{1}{c}{\Large{0.860}} &  \multicolumn{1}{c}{{\Large{0.855}}}  &  \multicolumn{1}{c}{{\Large{0.860}}}    &    \multicolumn{1}{c}{{\Large{0.848}}}      &   \multicolumn{1}{c}{{\Large{0.849}}}     &   \multicolumn{1}{c}{{\Large{0.866}}}   &   \multicolumn{1}{c}{{\Large{0.870}}}   &   \multicolumn{1}{c}{{\Large{0.871}}}   &   \multicolumn{1}{c|}{{\Large{0.877}}}   &   \multicolumn{1}{c}{{\Large{0.855}}}    \\
& $S_m\uparrow$        &   \multicolumn{1}{c}{\Large{0.864}}    & \multicolumn{1}{c}{\Large{0.866}} &\multicolumn{1}{c}{\Large{0.883}}  &  \multicolumn{1}{c}{\Large{0.892}}      &  \multicolumn{1}{c}{\Large{0.891}}   &  \multicolumn{1}{c}{\Large{0.894}}   &  \multicolumn{1}{c}{\Large{0.896}}    &  \multicolumn{1}{c}{\Large{0.892}}    &  \multicolumn{1}{c}{{\Large{0.896}}}&  \multicolumn{1}{c}{\Large{0.902}}    &  \multicolumn{1}{c}{{\Large{0.901}}}     &    \multicolumn{1}{c}{{\Large{0.895}}}     &   \multicolumn{1}{c}{\Large{0.899}}   &   \multicolumn{1}{c}{\Large{0.894}}&   \multicolumn{1}{c}{\Large{0.900}}&   \multicolumn{1}{c}{\Large{0.905}}&   \multicolumn{1}{c|}{\Large{0.906}}&   \multicolumn{1}{c}{\Large{0.890}}    \\
& $E_m\uparrow$     &   \multicolumn{1}{c}{\Large{0.905}}    & \multicolumn{1}{c}{\Large{0.917}}  &\multicolumn{1}{c}{\Large{0.911}}  &  \multicolumn{1}{c}{\Large{0.924}}      &  \multicolumn{1}{c}{\Large{0.927}}    &  \multicolumn{1}{c}{\Large{0.924}}    &  \multicolumn{1}{c}{\Large{0.927}}     &  \multicolumn{1}{c}{\Large{0.930}}  &  \multicolumn{1}{c}{{\Large{0.936}}}  &  \multicolumn{1}{c}{{\Large{0.933}}}  &  \multicolumn{1}{c}{{\Large{0.934}}}          &    \multicolumn{1}{c}{{\Large{0.929}}}    &   \multicolumn{1}{c}{{\Large{0.933}}}    &   \multicolumn{1}{c}{{\Large{0.935}}} &   \multicolumn{1}{c}{{\Large{0.940}}} &   \multicolumn{1}{c}{{\Large{0.936}}} &   \multicolumn{1}{c|}{{\Large{0.941}}} &   \multicolumn{1}{c}{{\Large{0.929}}}    \\
&$\mathcal{M}\downarrow$  &   \multicolumn{1}{c}{\Large{0.062}}    & \multicolumn{1}{c}{\Large{0.050}}  &\multicolumn{1}{c}{\Large{0.055}}  &  \multicolumn{1}{c}{\Large{0.049}}      &  \multicolumn{1}{c}{\Large{0.047}}  &  \multicolumn{1}{c}{\Large{0.048}}    &  \multicolumn{1}{c}{\Large{0.045}}    &  \multicolumn{1}{c}{\Large{0.044}}    &  \multicolumn{1}{c}{\Large{0.041}} &  \multicolumn{1}{c}{{\Large{0.042}}}  &  \multicolumn{1}{c}{{\Large{0.040}}}      &    \multicolumn{1}{c}{{\Large{0.044}}}     &   \multicolumn{1}{c}{{\Large{0.043}}}    &   \multicolumn{1}{c}{{\Large{0.041}}}   &   \multicolumn{1}{c}{{\Large{0.039}}}   &   \multicolumn{1}{c}{{\Large{0.039}}}   &   \multicolumn{1}{c|}{{\Large{0.037}}}   &   \multicolumn{1}{c}{{\Large{0.045}}}     \\
\bottomrule[1pt]
\bottomrule[2pt]
\rowcolor{mygray}
&$Ave-Rank$   &  
\multicolumn{1}{c}{\textbf{\Large{18}}}    & \multicolumn{1}{c}{\textbf{\Large{17}}}  &\multicolumn{1}{c}{\textbf{\Large{16}}}  &  \multicolumn{1}{c}{\textbf{\Large{15}}}      &  \multicolumn{1}{c}{\textbf{\Large{14}}}    &  \multicolumn{1}{c}{\textbf{\Large{13}}}  &  \multicolumn{1}{c}{\textbf{\Large{12}}} &  \multicolumn{1}{c}{\textbf{\Large{11}}}    &  \multicolumn{1}{c}{\textbf{\Large{7}}} &  \multicolumn{1}{c}{\textbf{\Large{6}}} &  \multicolumn{1}{c}{\textbf{\Large{4}}}    &    \multicolumn{1}{c}{\textbf{\Large{10}}}      &   \multicolumn{1}{c}{\textbf{\Large{8}}}   &   \multicolumn{1}{c}{\textbf{\Large{5}}} &   \multicolumn{1}{c}{\textbf{\Large{3}}} &   \multicolumn{1}{c}{\textbf{\Large{2}}} &   \multicolumn{1}{c|}{\textbf{\Large{1}}} &   \multicolumn{1}{c}{\textbf{\Large{9}}}   \\
\bottomrule[1pt]
\bottomrule[2pt]
  \end{tabular}
}
 	\setlength{\abovecaptionskip}{2pt}
    \caption{Quantitative comparison of different RGB-D SOD methods. $\uparrow$ and $ \downarrow$ indicate that the larger scores and the smaller ones  are better, respectively. The best three scores are highlighted in \textbf{bold}. The subscript in each model name is the publication year.  The corresponding used backbone network is listed below each model name.}
	\vspace{-5mm}
	\label{tab:Table1}
 \end{table*}

  \begin{figure*}
  \centering
   \setlength{\abovecaptionskip}{2pt}
  \includegraphics[width=0.9\textwidth]{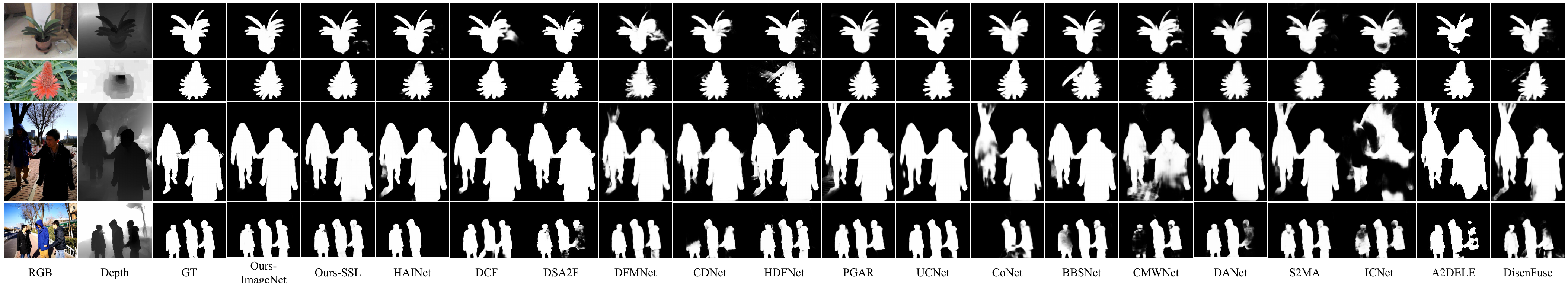}
  \caption{Visual comparison of different RGB-D SOD methods.}\label{fig:Figure6}
  \vspace{-3.5mm}
  \end{figure*}
  
   \begin{table}
    \centering
   
    \small
    \resizebox{\columnwidth}{!}
    {
	\begin{tabular}{c|ccccc|ccccc|c}
	\specialrule{1pt}{1pt}{1pt}
	No.& B$_i$ & + CM$_{jc}$ & + CM$_{jd}$ & + CL$_{jc}$ & + CL$_{jd}$ & $F_{\beta}^{max}$ & $F_{\beta}^{w}$ &$S_m $
	&$E_m $ &$\mathcal{M}$ & Params (MB)\\ \hline
	\specialrule{1pt}{1pt}{1pt}
	Model 1 & \checkmark & & & & 
	&  0.893 &0.829& 0.868& 0.904& 0.050 &36.50  \\
	\hline
	Model 2 & \checkmark & \checkmark & & & 	
	&  0.903 &0.846& 0.883& 0.921& 0.045&47.50 \\
	\hline
	
	Model 3 & \checkmark &\checkmark &\checkmark & & 
    &0.908 &	0.854 & 0.891& 0.927 &	0.042&58.49   \\
	\hline
	Model 4 & \checkmark & & &\checkmark & 	
	& 0.907 &0.854 &0.888&0.926&0.042&45.91 \\
	\hline
	Model 5 & \checkmark & & &\checkmark &\checkmark
	& 0.910&0.861&0.894	&0.931	& 0.040&52.18  \\
	\hline
	\specialrule{1pt}{1pt}{1pt}
	\rowcolor{mygray}
	Model 6 & \checkmark &\checkmark &\checkmark &\checkmark & \checkmark
	& \textBC{red}{0.920}	&\textBC{red}{0.877}&\textBC{red}{0.906}&\textBC{red}{0.941}&\textBC{red}{0.037}&74.17  \\
	\hline
	Model 7 & \checkmark & & & & 
	&  0.893 &0.829& 0.869& 0.905& 0.050 &78.15  \\
	\hline
	\end{tabular}
	}
	 \setlength{\abovecaptionskip}{2pt}
       \caption{Ablation experiments of the CDA module. B$_{i}$: Baseline with the ImageNet-pretrained backbone. + CM$_{jc}$ and + CL$_{jc}$: Using the joint consistency features in the cross-modal and cross-level fusion, respectively. + CM$_{jd}$ and + CL$_{jd}$: Using the joint difference features in the cross-modal and cross-level fusion, respectively. }
	\vspace{-8mm}
	\label{table:Table2}
\end{table}

\textbf{Effectiveness of Consistency-Difference Aggregation.} 
In  Tab.~\ref{table:Table2},  we show the performance contributed by different structures on all RGB-D SOD datasets in terms of the weighted average metrics ``Ave-Metric''. The baseline (Model $1$) is a two-stream FPN structure with deep supervision. We can see that this baseline has been able to suppress the A2DELE and DisenFuse. Based on this strong baseline, the performance gain is more convincing. Model $2$ vs. Model $1$ and Model $4$ vs. Model $1$ show the effectiveness of the jointly consistent features (JC) in cross-modal and cross-level integration, respectively. The JC can significantly improve the performance. Similarly, Model $3$ vs. Model $2$ and Model $5$ vs. Model $4$ show the advantages of the jointly differential features (JD). Model $6$ is equipped with the CDA module in both the cross-modal and cross-level fusion. Model $6$ vs. Model $3$ and Model $6$ vs. Model $5$ demonstrate the generalization of the CDA in the two kinds of information fusion, which complement each other without repulsion. 
To further evaluate the rationality of this module, the CDA is replaced with some operations of addition and convolution and a similar number of parameters are preserved. This new network is noted as Model $7$. Compared to it, Model $6$ has obvious advantages on performance.

We visualize intermediate features in the CDA module as shown in Fig.~\ref{fig:Figure7}. By explicitly superposing the influence of the cross-modal (or cross-level) joint consistency $F_{JC}$ and joint difference $F_{JD}$ under the constraint of saliency map,
the fused feature  $F_{JC,JD}^{RGB,D}$ (or $F_{JC,JD}^{4,3}$) can highlight salient regions well. Thus, the network equipped with the CDAs can more completely extract salient objects, whereas the network without the CDA is disturbed by a large number of non-salient areas (see the third row in Fig.~\ref{fig:Figure7}).  More comparisons are shown in the supplementary material.
\begin{table}[t]
    \centering
    
    \small
    \resizebox{\columnwidth}{!}
    {
	\begin{tabular}{c|ccccc|ccccc}
	\specialrule{1pt}{1pt}{1pt}
	No.& B$_r$ & + P1 & + P2 & + CM & + CL & $F_{\beta}^{max}$ & $F_{\beta}^{w}$ &$S_m $
	&$E_m $ &$\mathcal{M}$ \\ \hline
	\specialrule{1pt}{1pt}{1pt}
	Model 1 & \checkmark & & & & 
	&0.848 	&0.739  &0.841 	 &0.876  &	0.079  \\
	\hline
	Model 2 & \checkmark & \checkmark & & & 	
	&0.869 &	0.779	 &0.854 &	0.894 &	0.065 \\
	\hline
	Model 3 & \checkmark &\checkmark &\checkmark & & 
&	0.879 &	0.800 &	0.860 &	0.902& 	0.063   \\
	\hline
	Model 4 & \checkmark &\checkmark & &\checkmark & 	
 &	0.883 &	0.812 &	0.863 &	0.910 &	0.060  \\
	\hline
	Model 5 & \checkmark &\checkmark &\checkmark &\checkmark &
&	0.891  &	0.825 &	0.870 	&0.916 &	0.055  \\
	\hline
	Model 6 & \checkmark &\checkmark & & & \checkmark
&0.888 &	0.820  &	0.868 &	0.913 &	0.057  \\
	\hline
	Model 7 & \checkmark &\checkmark & \checkmark & & \checkmark
&	0.895 &	0.837 &	0.876 &	0.919 &	0.052 \\
	\hline
	\specialrule{1pt}{1pt}{1pt}
	Model 8 & \checkmark & & & \checkmark & \checkmark
 &	0.879 &0.792  &	0.868  &	0.904   &	0.062  \\
	\hline
	\rowcolor{mygray}
	Model 9 & \checkmark &\checkmark & \checkmark &\checkmark &\checkmark
	&\textBC{red}{0.906}&\textBC{red}{0.855}&\textBC{red}{0.890}&\textBC{red}{0.929}&\textBC{red}{0.045} \\
	\hline
	\end{tabular}
	}
	\setlength{\abovecaptionskip}{2pt}
    \caption{Ablation experiments of the SSL pretext tasks. All the results are based on the RGB-D SOD network. B$_{r}$: Baseline with the randomly initialized backbone. +P1: Initializing the encoder pretrained in pretext 1 (cross-modal auto-encoder).  +P2: Initializing the cross-modal layer and the decoder pretrained in pretext 2 (depth-contour estimation). + CM and + CL: Using the CDA in the cross-modal and cross-level fusion, respectively. }
	\vspace{-8mm}
	\label{table:Table3}
\end{table}

\begin{table}[t]
    \centering
   
    \small
    \resizebox{\columnwidth}{!}
    {
	\begin{tabular}{cccccc|ccccc}
	\specialrule{1pt}{1pt}{1pt}
	\multicolumn{6}{c|}{Model}& $F_{\beta}^{max}$ & $F_{\beta}^{w}$ &$S_m $
	&$E_m $ &$\mathcal{M}$ \\ \hline
	\specialrule{1pt}{1pt}{1pt}
	\multicolumn{6}{c|}{HAINet (Random Initialization)} 
   &0.875  &0.793  &0.857  &0.896  &0.064  \\
\multicolumn{6}{c|}{HAINet (Load our SSL-P1  pretraining weight)} 
&0.900 &	0.839 &	0.883& 	0.918 &	0.050 \\
\multicolumn{6}{c|}{HAINet (Load ImageNet pretraining weight)}
&0.920 	&0.871 	&0.905 &	0.936 &	0.039  \\
\hline \hline
	\multicolumn{6}{c|}{HDFNet (Random Initialization)} 
&  0.876 &	0.776 &	0.860 &	0.896& 	0.065  \\
\multicolumn{6}{c|}{HDFNet (Load our SSL-P1  pretraining weight)} 
&0.898 &	0.822 &	0.881 &	0.915 &	0.053  \\
\multicolumn{6}{c|}{HDFNet (Load ImageNet pretraining weight)}
 &0.916  &	0.860  &	0.901  &	0.934  &	0.040 
 \\
	\hline
	\end{tabular}
	}
	 \setlength{\abovecaptionskip}{2pt}
    \caption{Evaluation of existing methods with different initialization in terms of Ave-Metric.}
	\vspace{-5mm}
	\label{table:Table4}
\end{table}

\begin{table}[t]
    \centering
    
    \small
    \resizebox{\columnwidth}{!}
    {
	\begin{tabular}{cccccc|ccccc|c}
	\specialrule{1pt}{1pt}{1pt}
	\multicolumn{6}{c|}{Model}& $F_{\beta}^{max}$ & $F_{\beta}^{w}$ &$S_m $
	&$E_m $ &$\mathcal{M}$ & $Ave-Rank$\\ \hline
	\specialrule{1pt}{1pt}{1pt}
	\multicolumn{6}{c|}{Random Initialization} 
   & 0.879 & 	0.792 & 	0.868 & 	0.904&  	0.062 
  & 17  \\
\multicolumn{6}{c|}{NJUD + NLPR $\sim$2,000} 
&0.900 &	0.844 &	0.882 &	0.923&0.048 & 14\\
\multicolumn{6}{c|}{NJUD + NLPR + DUTLF-V2 (part) $\sim$4,000}
& 0.903 & 	0.848 & 	0.886 & 	0.926 & 	0.047& 12 \\
\multicolumn{6}{c|}{NJUD + NLPR + DUTLF-V2 $\sim$6,000}
 & 0.906 & 	0.855 & 	0.890&  	0.929 & 	0.045     & 9 \\
	\hline
	\end{tabular}
	}
	\setlength{\abovecaptionskip}{2pt}
    \caption{Evaluation of the SSL-based network with different number of pretraining examples in terms of Ave-Metric.}
	\vspace{-7mm}
	\label{table:Table5}
\end{table}


\textbf{Effectiveness of Self-supervised Pretraining.}
In Tab.~\ref{table:Table3}, we evaluate the effectiveness of the proposed SSL pretexts on the RGB-D SOD task in terms of the weighted average metrics ``Ave-Metric''. 
Firstly, we train the first pretext (P$1$), cross-modal auto-encoder. As shown in Fig.~\ref{fig:Figure8}, the predicted depth maps are close to the GT and the reconstructed RGB images clearly fit the semantics of the original images, such as the sky becomes blue and even the cat is colored with white. Next, we load the pretrained parameters from P$1$ to jointly train the second pretext (P$2$), depth-contour estimation. The visual results are shown in Fig.~\ref{fig:Figure8}. Model $2$ vs. Model $1$ shows that the P$1$ task can significantly improve the representation capability of the two-stream encoder. 
Model $3$ vs. Model $2$, Model $5$ vs. Model $4$ and Model $7$ vs. Model $6$ demonstrate the effectiveness of the P$2$ task.
In addition, Model $4$ vs. Model $2$ and Model $5$ vs. Model $3$ can further verify the effect of the CDA in the cross-modal fusion. While Model $6$ vs. Model $2$ and Model $7$ vs. Model $3$ show its benefits in the cross-level fusion. Finally, Model $9$ vs. Model $8$ indicates the overall contribution of two self-supervised learning pretexts. It can be seen that the performance is significantly improved with the gain of $7.95\%$ and $27.42\%$ in terms of the weighted F-measure and MAE, respectively. Besides, we load the pretraining weights of our proposed universal two-stream self-supervised encoder (SSL-P1) into the second and fourth methods (the third one is based on the ResNet-18 backbone) in Tab.~\ref{tab:Table1} as the initialization and retrain them. In Tab.~\ref{table:Table4}, we can see that using self-supervised pretraining weights can help two top-performing methods to consistently obtain a huge performance improvement compared to random initialization. This fully verifies the generalization ability of our SSL design. More comparisons can be found in the supplementary material.

\textbf{Evaluation of The Scale of Pretraining Data.} In Tab.~\ref{table:Table5}, we list the performance of using different numbers of SSL training images. We successively add $2,185$ training samples of NJUD and NLPR, $2,000$ samples of DUFLF-V2 and the rest $2,207$ ones of DUFLF-V2  to the training set. We train three SSL networks on $\sim$$2,000$ $-$ $\sim$$6,000$ unlabeled samples, and then load their parameters to initialize the downstream network, respectively. We can see that when only $\sim$$2,000$ samples are used, the SSL-based network already significantly outperforms the randomly initialized network.  With the increasing of training data, the performance is steadily improved, which shows that our SSL model has great potential.  More comparisons can be found in the supplementary material.

\begin{figure}
\includegraphics[width=0.78\linewidth]{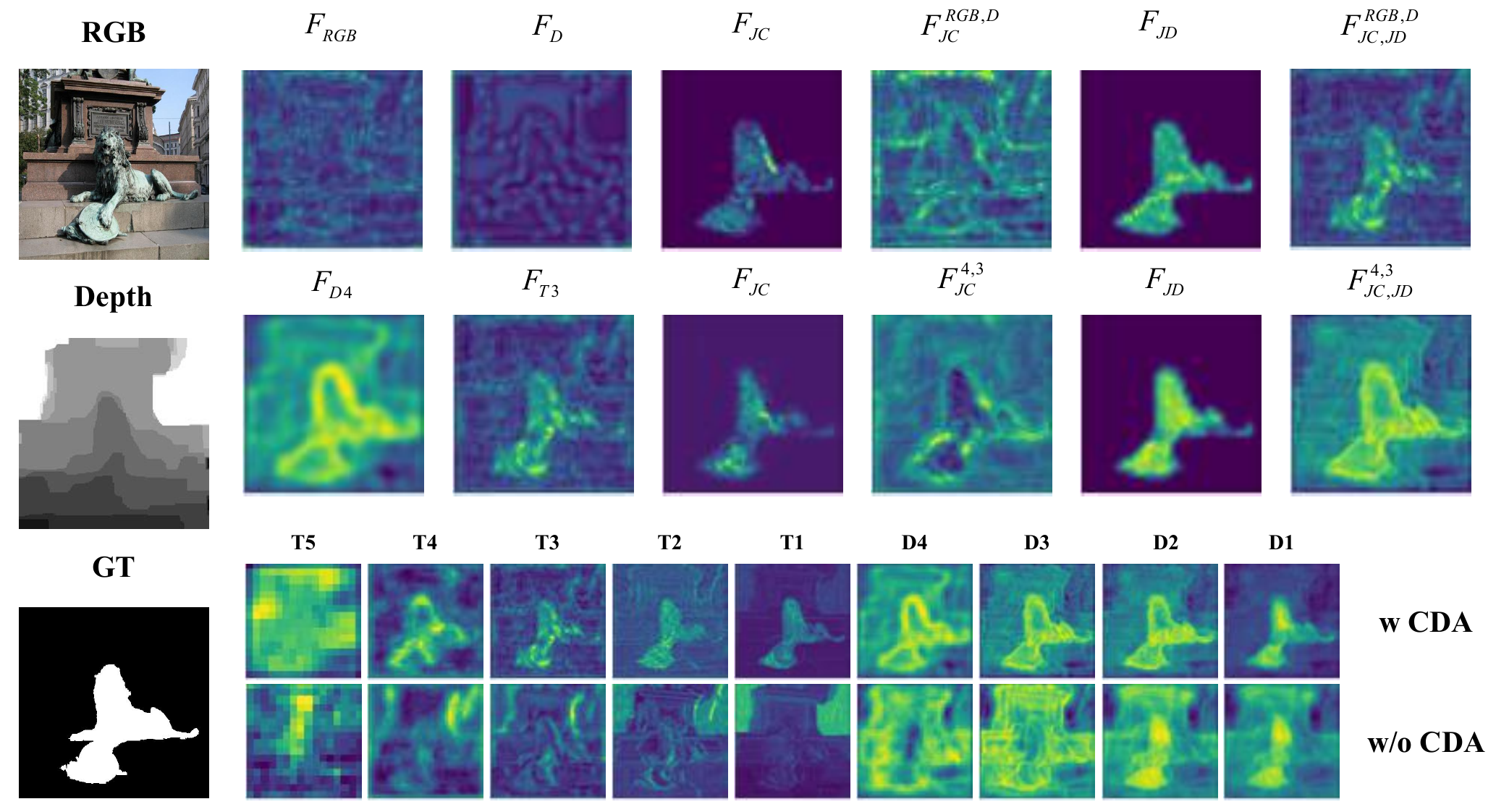}\\
        \centering
        \setlength{\abovecaptionskip}{-10pt}
        \caption{Illustration of the benefits of using the CDA. The first two rows correspond to different intermediate features in Fig.~\ref{fig:Figure4} in the cross-modal fusion and cross-level fusion, respectively. The third row shows visual comparison between the features of each level w and w/o CDA.}
\label{fig:Figure7}
\vspace{-3.5mm}
\end{figure}

\begin{figure}
\includegraphics[width=0.80\linewidth]{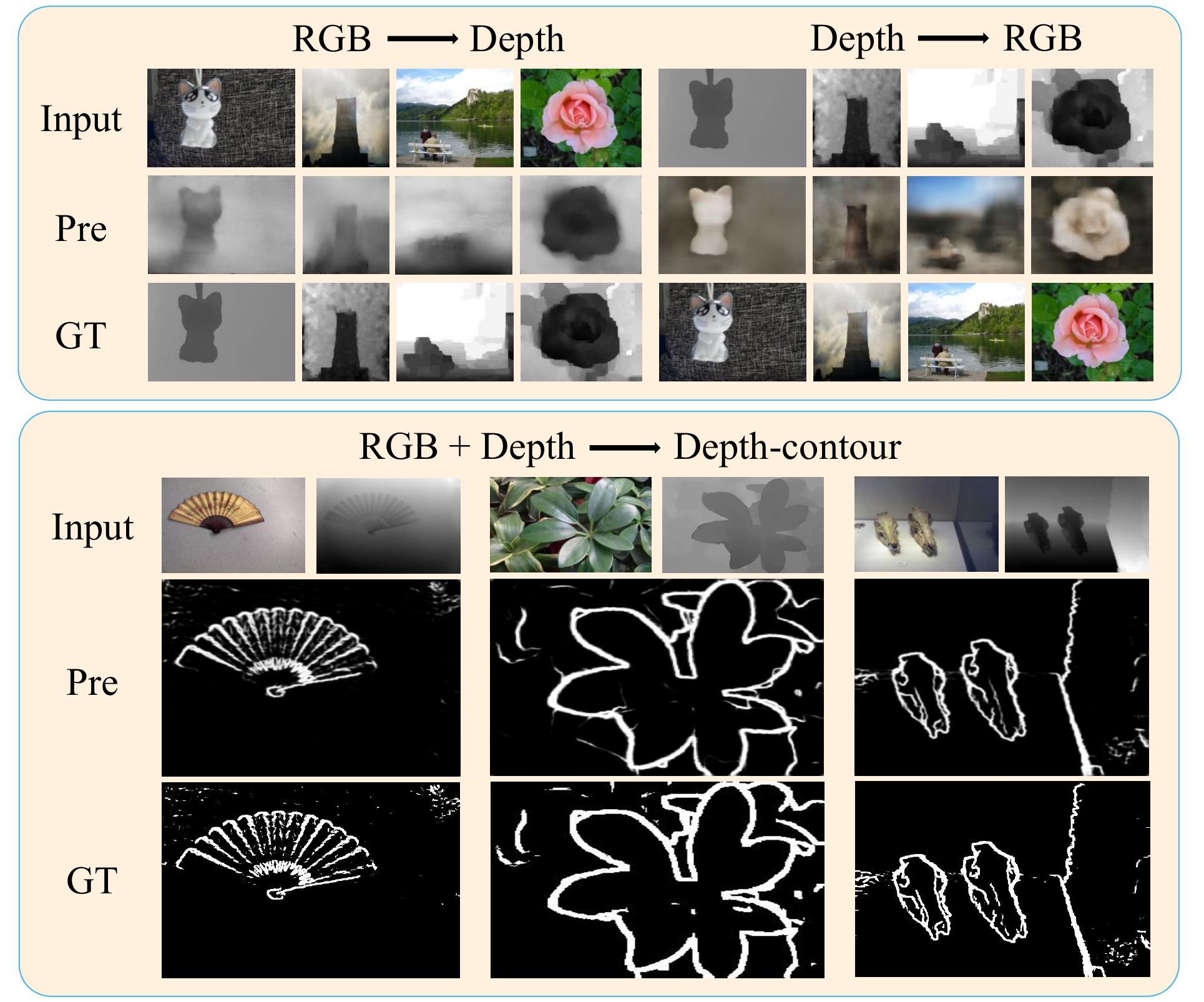}\\
        \centering
      \setlength{\abovecaptionskip}{-10pt}
        \caption{Visual results of the pretext tasks: cross-modal auto-encoder and depth-contour estimation.}
\label{fig:Figure8}
\vspace{-7mm}
\end{figure}

\section{Conclusion}
In this work, we propose a novel self-supervised learning (SSL) scheme to accomplish effective pretraining for RGB-D SOD without requiring human annotation. 
The SSL pretext tasks contain cross-modal auto-encoder and depth-contour estimation, by which the network can capture rich context and reduce the gap between  modalities. 
Besides, we design a consistency-difference aggregation module to combine cross-modal and cross-level information. 
Extensive experiments show our model performs well on RGB-D SOD datasets. As the first method of SSL in RGB-D SOD, it can be taken as a new baseline for future research.
\section{Acknowledgement}
This work was supported in part by the National Key R\&D Program of China \#2018AAA0102001, the National Natural Science Foundation of China \#61876202 and \#61829102, the Dalian Science and Technology Innovation Foundation \#2019J12GX039, the Liaoning Natural Science Foundation \#2021-KF-12-10, and the Fundamental Research Funds for the Central Universities \#DUT20ZD212. 
\bibliography{aaai22}
\bibliographystyle{aaai}
\bigskip

\end{document}